\documentclass[lettersize,journal]{IEEEtran}
\usepackage[utf8]{inputenc} 
\usepackage[T1]{fontenc}    
\usepackage{hyperref}       
\usepackage{url}            
\usepackage{booktabs}       
\usepackage{amsfonts}       
\usepackage{nicefrac}       
\usepackage{microtype}      
\usepackage{xcolor}         
\usepackage{hyperref}
\usepackage{url}
\usepackage{amsmath}
\usepackage{amssymb}
\usepackage{amsthm}
\usepackage{amsfonts}
\usepackage{bm}
\usepackage{graphicx}
\usepackage{caption}
\usepackage{subcaption}
\usepackage{thmtools}
\usepackage{thm-restate}
\usepackage{cleveref}
\usepackage{enumerate}
\usepackage{verbatim}
\usepackage{cite}
\usepackage[table,xcdraw]{xcolor}
\usepackage{wrapfig}
\usepackage{mhchem}
\usepackage{cleveref}
\usepackage{enumitem}
\usepackage{booktabs}
\usepackage{array}
\usepackage{float}
\usepackage{setspace}
\usepackage{tocloft}
\usepackage{amsmath,amssymb,amsfonts}
\usepackage{mathtools}

\theoremstyle{plain}
\newtheorem{theorem}{Theorem}[section]

\theoremstyle{definition}

\theoremstyle{remark}
\newtheorem{remark}[theorem]{Remark}

\graphicspath{{figures/}}

\begin{document}

\title{Exploiting Latent Linearity in LLMs Improves Explainable Molecular Representation Learning}

\author{
Zhuoran Li,
Xu Sun, 
Wanyu Lin$^{*}$,~\IEEEmembership{Member, IEEE},
Jiannong Cao,~\IEEEmembership{Fellow, IEEE}
\IEEEcompsocitemizethanks{\IEEEcompsocthanksitem
Zhuoran Li is with the Department of Computing, The Hong Kong Polytechnic University, Hong Kong SAR, P.R. China (e-mail: zhuor5.li@connect.polyu.hk).

Xu Sun is with the Department of Computing, The Hong Kong Polytechnic University, Hong Kong SAR, P.R. China (e-mail: yeluo.sun@connect.polyu.hk).

Wanyu Lin is with the Department of Computing and Department of Data Science and Artificial Intelligence, The Hong Kong Polytechnic University, Hong Kong SAR, P.R. China (e-mail: wan-yu.lin@polyu.edu.hk).

Jiannong Cao is with the Department of Computing, The Hong Kong Polytechnic University, Hong Kong SAR, P.R. China (e-mail: jiannong.cao@polyu.edu.hk).

\textit{(*Corresponding author: Wanyu Lin.)} 

	}	
}

\markboth{Journal of \LaTeX\ Class Files,~Vol.~14, No.~8, August~2021}%
{Shell \MakeLowercase{\textit{et al.}}: A Sample Article Using IEEEtran.cls for IEEE Journals}

\IEEEpubid{0000--0000/00\$00.00~\copyright~2021 IEEE}

\maketitle

\begin{abstract}

Large language models (LLMs) have demonstrated broad utility across molecular domains, spanning drug discovery and materials design. Analyzing LLMs' latent representations is crucial for elucidating their underlying mechanisms, improving explainability, and ultimately advancing downstream performance. We propose \textit{MoleX}, a simple yet effective framework that decomposes molecular embeddings within LLM representations into a concept‑aligned space for explainable molecular representation learning. We further show that these high‑dimensional embeddings admit a linear mapping onto chemically consistent concepts. Our analysis suggests that the uncovered linearity aligns with established chemical principles, indicating a mechanistically explainable latent structure in LLM representations for scientific applications. When applied to downstream tasks, this latent linearity improves both predictive and explanatory performance. Extensive experiments demonstrate that \textit{MoleX} outperforms existing approaches in accuracy, explainability, and efficiency---achieving CPU inference on large-scale datasets 300$\times$ faster with 100,000 fewer parameters than LLMs. 

\end{abstract}

\begin{IEEEkeywords}
Molecular representation learning, explainability, molecular property prediction, large language models.
\end{IEEEkeywords}

\section{Introduction}
\IEEEPARstart{M}{olecular} property prediction exemplifies a fundamental scientific task that infers molecular properties from given structures, serving as a cornerstone in fields such as computational chemistry and biology~\cite{xia2024understanding, yang2019analyzing, yang2025curriculumaware, zhou2025clmfap, zhong2025automatic, yeyunchen2025mipt, rao2025quadruple}. Modern LLMs have achieved exceptional predictive performance by learning expressive molecular representations that capture chemical semantics from large textual corpora~\cite{chithrananda2020chemberta, ahmad2022chemberta, li2024towards, feng2024generation, li2025geometry, liu2025multimodal, zheng2025large}. However, LLMs remain mechanistically opaque due to their black-box nature~\cite{wu2023black, kang2024quantitative, sharkey2025open}: \textit{why do these representations work, and can we trust what they learn}? Though prior efforts identify key substructures linked to properties, few derive them from the model’s representational mechanisms, reducing explanation faithfulness. Among them, description-based approaches~\cite{wellawatte2025human} are hindered by descriptive subjectivity and the complexity of molecular patterns~\cite{hase2020evaluating, kunz2024properties}, while attention-based methods~\cite{balaji2023gpt, zheng2024large} often fail to reflect true feature importance~\cite{bai2021attentions}. Notably, \textit{Lamole}~\cite{wanyu-nips24} provides a state-of-the-art solution that delivers both accurate and explainable predictions, but it remains highly model-specific~\cite{jain2019attention} and offers only local explanations~\cite{liu2022rethinking}. In summary, these studies do not adequately illuminate how molecular structures are encoded in latent space and how those encodings produce the predicted properties.

\begin{figure}[t] 
  \centering   
  \includegraphics[width=0.95\linewidth]{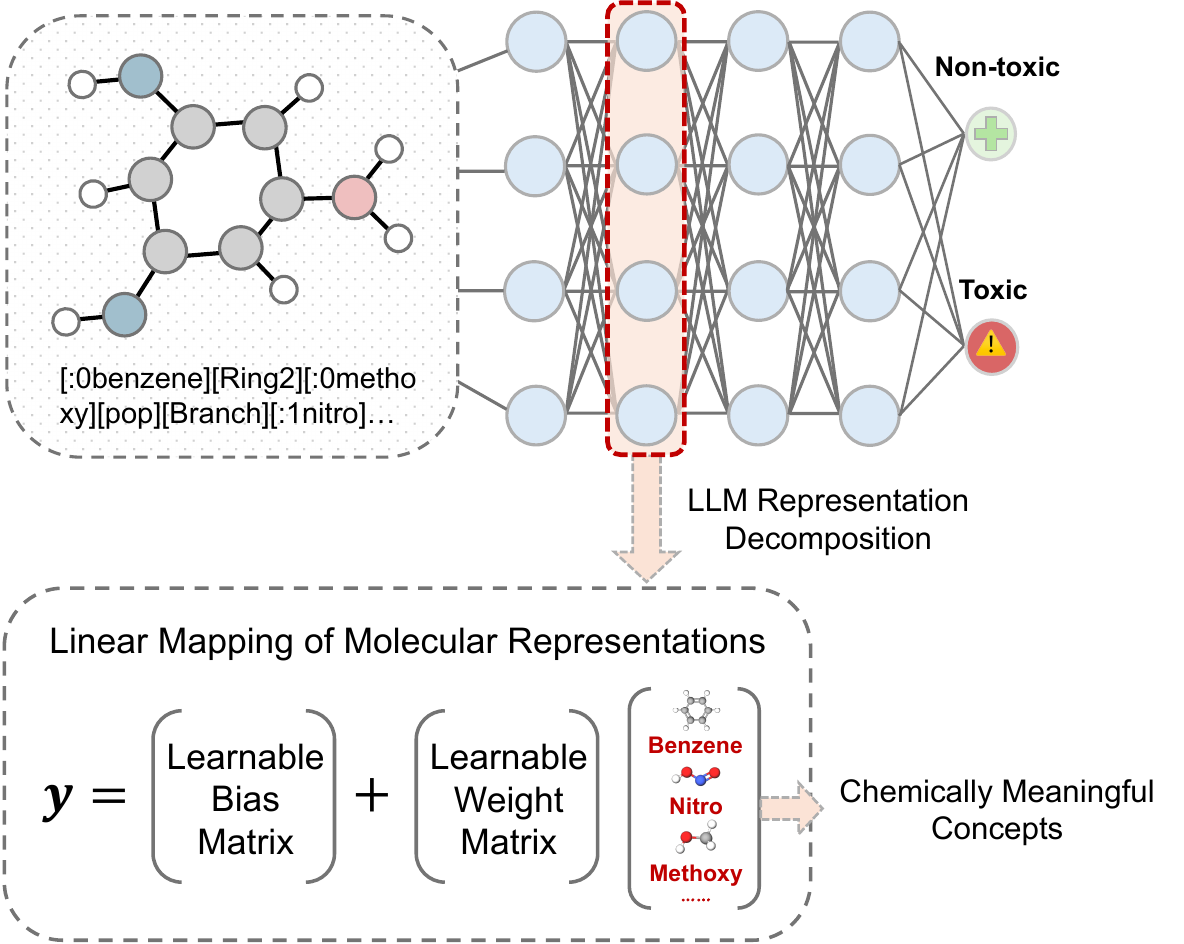}
  \caption{\centering The molecular representations are decomposed \\ into a linear mapping of functional groups.}
\label{fig:pipeline2}
\vspace{-4mm}
\end{figure}

To address the limitations, we exploit LLMs’ hidden representations to develop a mechanistically explainable approach that is robust to input sensitivity, model specificity, and interpretive subjectivity. Conceptually, LLM representations encode the model’s latent reasoning patterns for a given input (\textit{i.e.,} how the LLM uses its ``knowledge'' to reason)~\cite{ren2024representation, petroni2019language, he2024harnessing}, so understanding them is key to unlocking LLMs’ full potential in scientific tasks~\cite{fang2022geometry}. To achieve domain‑aligned explainability, we aim to map these representations to chemically meaningful concepts. Inspired by the linear representation hypothesis~\cite{park2024the, modell2025origins}, we investigate whether and how LLM‑derived molecular representations can be \textit{linearly decomposed} into chemical concepts (as shown \Cref{fig:pipeline2}). Although LLMs are built on complex nonlinear transformations, we posit that operative molecular semantics reside in an approximately linear geometry with respect to \textit{functional groups}, which are widely adopted concepts for property attribution in molecular science~\cite{bruice2017organic, bader1994theoretical, siddique2024exploring, gani2019group, yan2010oxidation}. In particular, we achieve this by feeding Group SELFIES~\cite{cheng2023group}, a textual representation that partitions molecules into functional group units (see Supplementary Material I for an visualized illustration), into the LLM to obtain functional-group–based tokens for our subsequent study.

\begin{figure*}[t]
    \centering
    \includegraphics[width=1.0\textwidth]{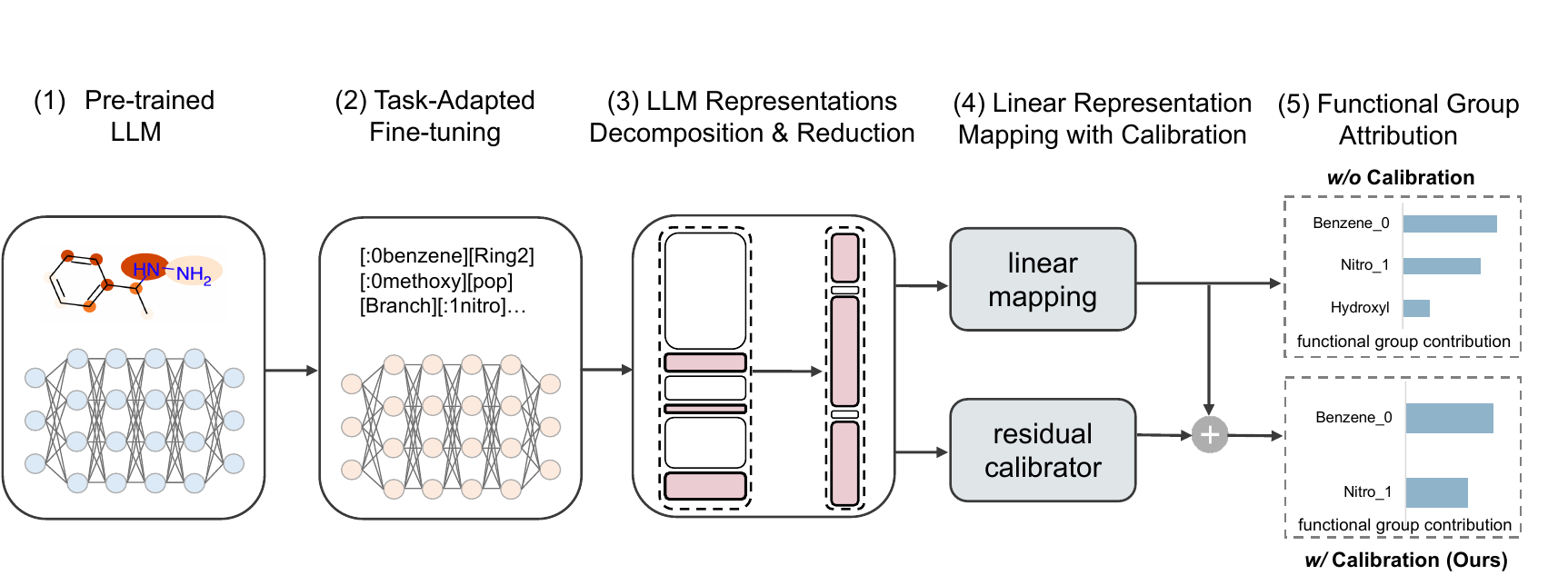}
    \caption{\textbf{The overview of {\em MoleX}}: (1) employ ChemBERTa-2 as the foundation model, (2) fine-tune it using a tailored objective to retain maximal task-relevant semantics, (3) decompose and project latent representations onto chemically meaningful concepts, (4) train a linear predictor on the derived concepts, (5) incorporate residual calibration to iteratively refine model predictions.}
    \label{fig:pipeline}
    \vspace{-4mm}
\end{figure*}

In light of this, we introduce \textit{MoleX} (outlined in \Cref{fig:pipeline}), which constructs a low-dimensional conceptual space spanned by functional group vectors and learns a sparse linear mapping from LLM-derived representations onto this space. Each molecule is represented as a weighted combination of concepts, with weights reflecting predictive significance. This yields concept-level explanations grounded in chemical principle while maintaining strong predictive performance. However, while LLMs encode rich molecular semantics, simple linear representations may fail to capture intricate patterns, leading to mispredictions on complex tasks. To address this, we design a residual calibration strategy that iteratively corrects prediction errors from the base linear predictor. Beyond improving predictive accuracy, residual calibration also increases explanation faithfulness by progressively probing molecular substructures that are chemically consistent with the ground truth. We summarize our main contributions as follows:

\newpage

\begin{enumerate}

    \item We propose a chemically consistent decomposition framework that reveals hidden linearity in LLMs' latent geometry, enhancing mechanistic explainability in molecular representation learning.

    \item We design a residual calibration strategy to compensate for the linear predictor’s limited expressiveness, improving both predictive accuracy and explanation faithfulness.

    \item We demonstrate that \textit{MoleX} achieves state-of-the-art accuracy, explainability, and efficiency on benchmark molecular property prediction tasks with over 100,000 fewer parameters than existing deep models.

    \item We present a theoretical analysis of how latent linearity boosts explainability, providing provable insights for molecular property prediction.
    
\end{enumerate}

Empirical evidence confirms that linear structure exists in LLMs’ representational geometry and can be expressed as a combination of chemical concepts. Relying solely on simple techniques, \textit{MoleX} provides meaningful insight into understanding complex LLM representations from the view of domain knowledge. Moreover, we introduce a residual calibration approach to capture errors from the base linear predictor, further improving both accuracy and explainability. Taken together, \textit{MoleX} demonstrates that latent linearity in LLMs provides a principled basis for effective and explainable molecular representation learning. When used for molecular property prediction, the exploited linearity notably improves predictive accuracy, explanation faithfulness, and computational efficiency, underscoring its applicability in practical scenarios.

\section{Related Work}\label{sec:related}

\textbf{Explainable Molecular Representation Learning.}  Molecules can naturally be modeled as graphs, so that many explainable GNNs have been proposed to capture structure–property relationships \cite{lin2021generative, pope2019explainability}. However, their explanations typically focus on atom- or bond-level importance, which often lack chemical consistency and fail to reflect higher-level functional group interactions. In contrast, LLMs have shown strong capabilities in learning substructure-level interactions from text-based molecular representations \cite{wanyu-nips24, ross2022large, chithrananda2020chemberta}. Despite their expressive power, LLMs remain opaque, raising concerns about faithfulness and reliability, especially in safety-critical domains such as drug design \cite{nips24-d4exp}. Our approach bridges this gap by aligning LLM-based chemical concepts with a transparent linear predictor, enabling both accurate predictions and chemically consistent explanations.

\textbf{Explainability Methods for LLMs.}  
To interpret LLMs, various explanation techniques have been proposed. Gradient-based methods estimate input importance via output derivatives \cite{sundararajan2017axiomatic}, but their sensitivity to perturbations limits robustness \cite{kindermans2019reliability, adebayo2018sanity}. Attention-based methods rely on attention weights \cite{hoover2020exbert}, yet these weights can not faithfully reflect model reasoning \cite{jain2019attention, serrano2019attention}. Perturbation-based methods analyze output changes under input modifications \cite{ribeiro2016should}, but often suffer from instability due to randomness in the perturbation process \cite{agarwal2021towards}. In contrast, our method directly decomposes LLM representations into interpretable chemical concepts and applies an explainable linear mapping for prediction, combining the expressive power of LLMs with the simplicity and faithfulness of linear predictor to produce accurate and explainable outputs.

\section{Preliminaries}\label{sec:preliminaries}

Let \( \mathcal{G} = \{ (g^{(i)}, y^{(i)}) \} \) be a dataset of molecular graphs \( g^{(i)} \) and their properties \( y^{(i)} \). Our goal is to predict \(y\) from molecular structure while providing explanations that remain aligned with chemically meaningful substructures, namely functional groups. We first convert each \( g^{(i)} \) into Group SELFIES, denoted as \( x^{(i)} = \{ x_1^{(i)}, \ldots, x_{j^{(i)}}^{(i)} \} \), where \( x_j^{(i)} \) is the \( j \)-th group. We fine‑tune an LLM \(f\) for molecular property prediction and obtain its representation \(f(x)\) (\textit{i.e.,} the latent representation of \(x\)). The overall predictor consists of an explainable linear predictor \(h\) and a residual calibrator \(r\). Concretely, we first compress \(f(x)\) into a low-dimensional principal component analysis (PCA) representation \(z(x)\), and split the PCA dimensions into two disjoint index sets \(I_H\) and \(I_R\) with \(I_H\cap I_R=\varnothing\). We denote the corresponding subvectors by \(f_H(x)\) and \(f_R(x)\) (in our implementation, \(f_H(x)=z_H(x)\) and \(f_R(x)=z_R(x)\)). This construction yields non-overlapping features for the linear predictor and the residual calibrator, so their effects are separable in the logit space.

Specifically, we first train \(h\) on \(f_H(x)\) and then freeze its parameters. Next, we train \(r\) on \(f_R(x)\) to correct the prediction errors of the base predictor \(h\) by minimizing:
\begin{align} \label{residualpred_objective}
\min_{r} \; \mathbb{E}_{(x, y) \sim \mathcal{D}} \left[ \mathcal{L}\left( \sigma\!\left(h\left( f_H(x) \right) + r\left( f_R(x) \right)\right),\, y \right) \right],
\end{align}

where \( \mathcal{D} \) is the training dataset, and \(h(\cdot)\) and \(r(\cdot)\) can be interpreted as logits. Building on \cite{sebastiani2002machine}, we use coefficients in the linear predictor to quantify feature effect on property $y$. Let \([f_H(x)]_j\) be the \(j\)-th explainable feature and let \(w_j\) be its coefficient in \(h\). Then the instance-level contribution score is computed as \(c_j \;=\; w_j\,[f_H(x)]_j\), which measures the contribution of that feature to the prediction under the linear predictor. Therefore, we can statistically quantify the contribution of the \(j\)-th functional group to property \(y\). For simplicity, we omit the superscript \( ^{(i)} \) in the following descriptions.

\section{Our Framework: {\em MoleX}}
\label{sec:methods}

The proposed approach leverages simple yet principled tools from statistical machine learning to exploit latent linearity within the intricate representational geometry of LLMs. Specifically, \emph{MoleX} consists of two stages: (1) decomposing high-dimensional molecular representations into chemical concepts, and (2) learning a concept-level linear mapping with residual calibration. Our design is motivated by a simple goal: preserve the expressive power of LLM representations, but uncover the task-relevant information through a small set of explainable features that can be statistically attributed.

In particular, we first fine-tune a pre-trained LLM using an information bottleneck-inspired objective to construct informative molecular representations. The resulting embeddings are then decomposed at the functional-group level, and we further apply dimensionality reduction to obtain a compact latent space suitable for linear modeling. A linear predictor \( h \) is trained on an explainable subset of these features to predict molecular properties, while a residual calibrator \( r \) corrects the remaining errors in logit space using the complementary feature subset. The final prediction subsequently combines both \( h \) and \( r \) to yield accurate and explainable results.

\subsection{Constructing and Decomposing LLM Representations}

\textbf{LLM Fine-Tuning.} To learn informative representations, we fine-tune the LLM on Group SELFIES, aligning input tokens with chemically consistent substructures. In our task, this step encourages the latent representation to emphasize chemically meaningful patterns (\textit{e.g.,} functional‑group‑level semantics) rather than spurious molecular fragments. Nevertheless, standard fine-tuning often entangles molecular semantics with irrelevant noise. To address this, we introduce a fine-tuning objective inspired by the variational information bottleneck (VIB)~\cite{alemi2022deep}. Intuitively, we want the latent representation \(e\) to maximally retain information predictive of the property \(y\) while discarding redundant details from the molecular structure \(x\) that do not support explainable structure–property reasoning.

Given a molecular input \( x \), property \( y \), and the LLM’s latent representation \( e \), we define \( p_0(e) \) as the prior over \( e \), and \( q_{\theta}(y \mid e) \) as a variational approximation to the conditional distribution of \( y \) given \( e \). We quantify the desired trade-off using mutual information: \(I(e;y)\) measures how predictive the representation is for the downstream property, while \(I(e;x)\) measures how much input information is preserved. The mutual information between \( e \) and \( y \) is given by:
\begin{align*}
I(e; y) = \textstyle \mathbb{E}_{p(e, y)}\left[ \log \frac{p(e, y)}{p(e)p(y)} \right] = \mathbb{E}_{p(e, y)}\left[ \log \frac{p(y \mid e)}{p(y)} \right], \nonumber
\end{align*}
and we introduce the variational decoder \( q_{\theta}(y \mid e) \) and optimize the tractable surrogate \( \mathbb{E}_{p_{\theta}(e \mid x)}\!\big[ -\log q_{\theta}(y \mid e) \big] \). The mutual information between \( e \) and \( x \) is defined as: 
\begin{align*}
I(e; x) &= \mathbb{E}_{p(e, x)}\!\left[ \log \frac{p(e \mid x)}{p(e)} \right] \\
        &= \mathbb{E}_{p(x)}\!\left[ D_{\mathrm{KL}}\!\left( p_{\theta}(e \mid x) \,\big\|\, p(e) \right) \right].
\end{align*}
Since the marginal distribution \( p_{\theta}(e) \) is intractable, we approximate it with the prior \( p_0(e) \). Under this approximation,
\[
\mathbb{E}_{p(x)}\!\left[ D_{\mathrm{KL}}\!\left( p_{\theta}(e \mid x) \,\big\|\, p_{0}(e) \right) \right]
= I(e; x) + D_{\mathrm{KL}}\!\big(p_{\theta}(e)\,\|\,p_0(e)\big)
\]
provides a tractable upper bound on \( I(e; x) \) up to an \( x \)-independent term, which allows us to control (\textit{i.e.,} an upper-bound) the mutual information between \( e \) and \( x \). Inspired by \cite{kingma2015variational}, we approximate encoder \( p_{\theta}(e \mid x) \) by a Gaussian distribution. Let \( f_e^{\mu}(x) \) and \( f_e^{\Sigma}(x) \) be neural networks that output the mean and covariance matrix of latent variable \( e \), then the encoder can be expressed as \(
p_{\theta}(e \mid x) = \mathcal{N}\left( e \,\big|\, f_e^{\mu}(x), f_e^{\Sigma}(x) \right).\)
Applying the reparameterization trick, we sample \( e \) as \(
e = f_e^{\mu}(x) + f_e^{\Sigma}(x)^{1/2} \cdot \epsilon, \text{where } \epsilon \sim \mathcal{N}(0, I).\) Eventually, we design the LLM fine-tuning objective as: 
\begin{equation}\label{ft_loss}
\begin{aligned}
\mathcal{L}(\theta)
  = \sum_{(x, y) \in \mathcal{S}_F} \Big(
      \mathbb{E}_{p_{\theta}(e \mid x)}\!\big[ -\log q_{\theta}(y \mid e) \big] + 
      \\ \beta \cdot D_{\mathrm{KL}}\!\big( p_{\theta}(e \mid x) \,\big\|\, p_{0}(e) \big)
    \Big),
\end{aligned}
\end{equation}
where \( \beta \) controls the trade-off between compression and performance, \( q_{\theta} \) serves as the decoder, and \( \mathcal{S}_F \) denotes the fine-tuning dataset. Specifically, the first term, \( \mathbb{E}_{p_{\theta}(e \mid x)} \big[ -\log q_{\theta}(y \mid e) \big] \), encourages the latent representation \( e \) to retain task-relevant information about \( y \). The second term, \( \beta \cdot D_{\mathrm{KL}} \big( p_{\theta}(e \mid x) \,\|\, p_0(e) \big) \), regularizes \( e \) toward the prior to limit information captured from \( x \) that is not necessary for predicting \( y \). In summary, this objective ensures that the fine-tuned LLM learns representations \( e \) that retain maximal task-relevant semantics from \( y \) while suppressing redundancy from \( x \). By constraining information flow, our method highlights chemically contributory information while filtering irrelevant content, thereby enhancing specificity and explainability for molecular representation learning. We formulate this guarantee as follows (demonstration provided in the Supplementary Material A).

\begin{remark} \label{theorem3}
Let \( \mathcal{L}(\theta) \) denote the loss in \Cref{ft_loss}. Minimizing \( \mathcal{L}(\theta) \) encourages the learned molecular representations \( e \) to balance predictive accuracy via the negative log-likelihood term and compression toward the prior via the KL regularization, which serves as an information-bottleneck- inspired inductive bias for molecular property prediction.
\end{remark}

\textbf{Representation Decomposition.} 
We begin by encoding each molecule using Group SELFIES, a structured molecular representation that preserves functional group boundaries by design. We train a functional group-level tokenizer that individually processes each token/functional group, which is then passed through the fine-tuned LLM to obtain a fixed-size embedding. This design reduces interference between tokens and enables clear attribution of learned features to their corresponding functional groups. Finally, to obtain a global representation, we aggregate the local embeddings of all functional groups within a molecule into a single vector. This vector retains the semantic expressiveness of LLM representations while preserving a decomposition aligned with chemical concepts. In essence, we decompose high‑dimensional molecular representations into functional group vectors and aim to explain each group’s contribution to the predicted property. We use UMAP~\cite{mcinnes2018umap} to visualize the decomposed functional group semantics in 3D space, as shown in \Cref{fig:umap} (more visualizations are provided in the Supplementary Material F).

\subsection{Learning a Linear Mapping with Residual Calibration}

\textbf{Dimensionality Reduction.} LLM-derived molecular embeddings are high-dimensional and contain redundant chemical semantics that can obscure information relevant to molecular property prediction. Therefore, dimensionality reduction is needed before learning a predictor. In this work, we leverage PCA as it provides a statistically explainable way to compress representations~\cite{abdi2010principal}. Specifically, PCA projects high-dimensional LLM embeddings onto orthogonal directions of maximal explained variance, producing compact representations in which we can later separate explainable and residual components for calibration.

Formally, let \(f(x)\in\mathbb{R}^d\) denote the LLM representations of molecule \(x\), and let \(\mathcal{S}_D\) be the dataset used to run the PCA. We define the empirical mean as \(\mu \coloneqq |\mathcal{S}_D|^{-1}\!\sum_{(x,y)\in\mathcal{S}_D} f(x)\), and the empirical covariance as:
\[
\Sigma \;\coloneqq\; |\mathcal{S}_D|^{-1}\!\sum_{(x,y)\in\mathcal{S}_D} \big(f(x)-\mu\big)\big(f(x)-\mu\big)^\top \;\in\; \mathbb{R}^{d\times d}.
\]
PCA aims to learn an orthonormal loading matrix \(U_k\in\mathbb{R}^{d\times k}\) such that \(U_k^\top U_k=I_k\) to maximize the retained variance:
\[
U_k \;=\; \arg\max_{U\in\mathbb{R}^{d\times k}\,:\,U^\top U=I_k}\; \mathrm{Tr }\!\big(U^\top \Sigma U\big),
\]
whose solution consists of the top-\(k\) eigenvectors of \(\Sigma\) with eigenvalues \(\lambda_1 \ge \cdots \ge \lambda_d\). The original molecular representation is then reduced by projection as follows:
\[
z(x) \;\coloneqq\; U_k^\top \big(f(x)-\mu\big)\in\mathbb{R}^k,
\]
so that \(z(x)\) provides a compact representation for downstream tasks. In our setting, these PCA coordinates serve as a denoised latent space where linear predictors can capture dominant structure--property signals. The dimension \(k\) is chosen by an explained‑variance threshold \(\sum_{i=1}^k \lambda_i\,/\,\sum_{i=1}^d \lambda_i \ge \tau\) or via validation on the molecular property \(y\) (\textit{i.e.,} we include an empirical study in \Cref{ablation} to illustrate how to determine the optimal \(k\)). The parameters \((\mu, U_k)\) are estimated once on \(\mathcal{S}_D\) and then fixed for subsequent linear mapping. Accordingly, we extract statistically relevant dimensions from noisy latent space. Noteworthily, after each functional‑group token is encoded by the LLM and then dimension‑reduced, we construct a mapping table linking each functional group to its corresponding reduced molecular embedding. This allows us to trace each feature back to a specific functional group by looking up this table during attribution with the linear predictor.

\textbf{Linear Mapping of Concepts.}  
We train a linear predictor on the explainable representation \(f_H(\tilde{x})\in\mathbb{R}^n\) for property prediction. The motivation is that, for molecular tasks, a linear mapping in a concept-aligned space provides a direct link between substructures and predicted properties, which is crucial for chemically grounded explanations. In our implementation, \(f_H(\tilde{x})\) corresponds to the explainable PCA subvector \(z_H(x)\) (thus \(n=|I_H|\)). It can be formulated as:
\[
h\big(f_H(\tilde{x})\big) \;=\; \sigma\!\left(w^\top f_H(\tilde{x}) + b\right),\quad w\in\mathbb{R}^{n},\; b\in\mathbb{R},
\]
where \(\sigma\) is the sigmoid function. The log‑odds then satisfy:
\[
\log\!\left(\frac{h(f_H(\tilde{x}))}{1-h(f_H(\tilde{x}))}\right)
\;=\; w^\top f_H(\tilde{x}) + b,
\]
so the marginal effect on log-odds with respect to the \(j\)-th dimension of \(f_H(\tilde{x})\) is
\[
\frac{\partial}{\partial\, [f_H(\tilde{x})]_j}
\log\!\left(\frac{h(f_H(\tilde{x}))}{1-h(f_H(\tilde{x}))}\right)
\;=\; w_j.
\]
Concretely, this interpretation follows the standard regression theory~\cite{hastie2009elements}. The coefficient \(w_j\) reflects the per-unit effect on the log-odds along the \(j\)-th feature, while the instance-level contribution is \(w_j\,[f_H(\tilde{x})]_j\). Consequently, we can attribute the \(j\)-th functional group’s contribution to the molecular property. This suggests an explainable view of how the model’s decision varies along explainable latent dimensions.

\begin{figure*}[t]
    \centering
    \begin{subfigure}[t]{0.58\textwidth}
        \includegraphics[width=\linewidth]{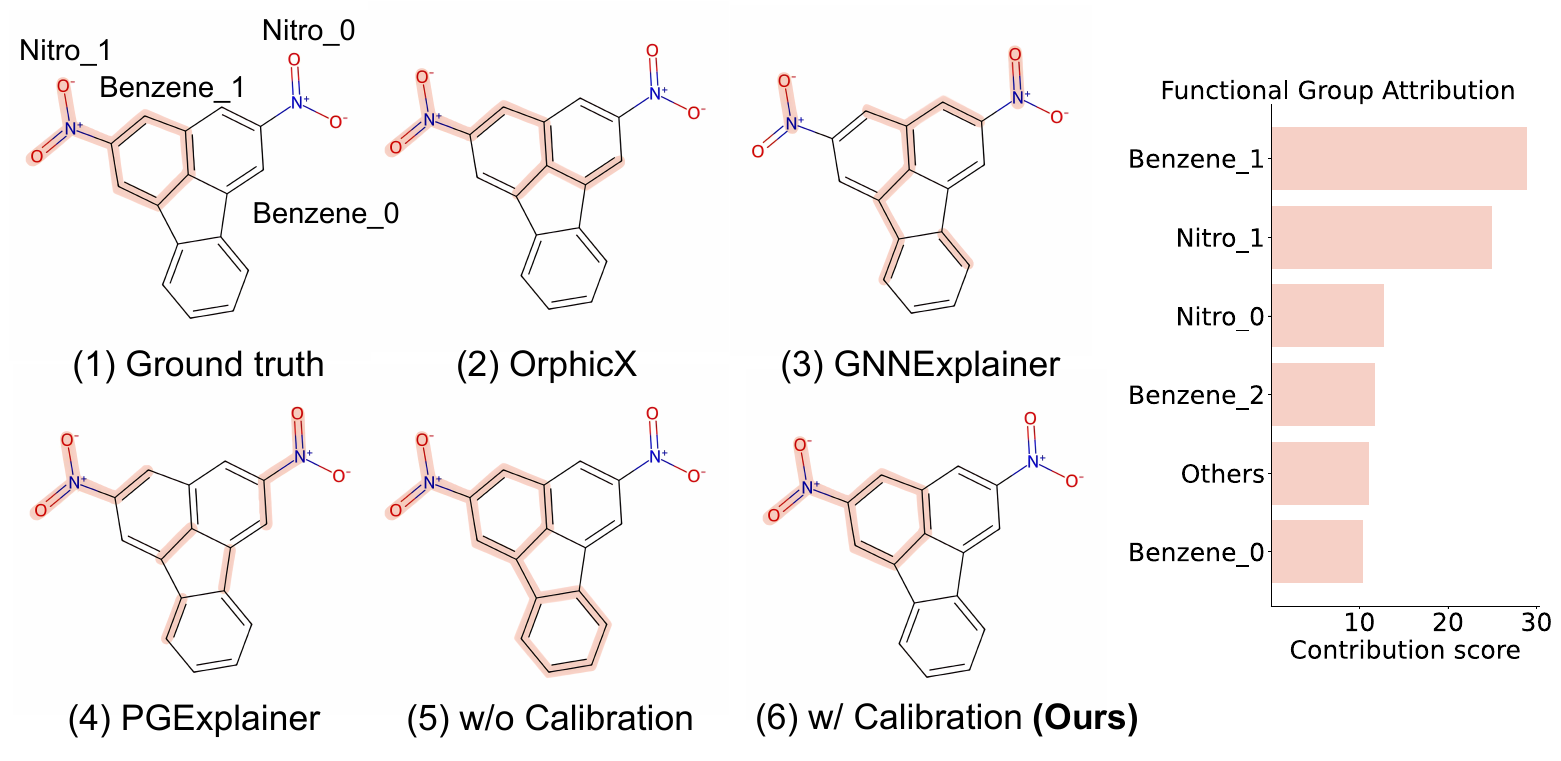}
        \label{fig:gt_mutag}
    \end{subfigure}
    \hfill
    \begin{subfigure}[t]{0.41\textwidth}
        \includegraphics[width=\linewidth]{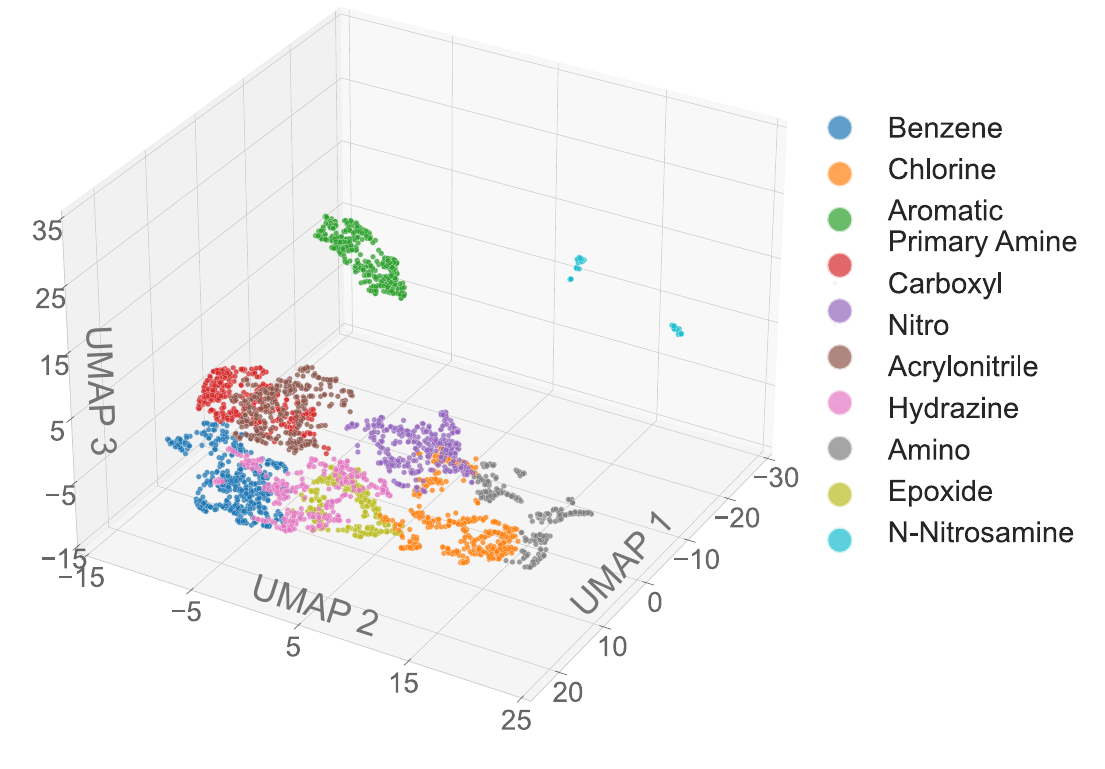}
        \label{fig:umap_mutag}
    \end{subfigure}
    \caption{Illustration of revealed functional groups. (\textbf{Left}) Explanation visualization across different methods, with functional group attribution offered by \textit{MoleX}. (\textbf{Right}) The 3D UMAP displays well-separated clusters for decomposed functional groups, indicating intrinsic linear structure in the LLM representations with concept-aligned separability. Rare groups form isolated clusters, while chemically related groups remain proximal yet distinguishable, reflecting linearly decodable chemical semantics.}
    \label{fig:umap}
    \vspace{-2mm}
\end{figure*}

\textbf{Residual Calibration.} 
The linear mapping \( h \), while highly explainable, is inherently limited in expressiveness due to its simple architecture, potentially underfitting complex patterns. We therefore introduce a residual calibrator \( r \) that learns an additive correction in logit space while keeping \(h\) fixed, so that the overall predictor can improve accuracy without obscuring the attribution of the base linear terms. Concretely, we first compute a PCA representation
\(z(x) \coloneqq U_k^\top\big(f(x)-\mu\big)\in\mathbb{R}^k\), and then split the PCA coordinates into two disjoint index sets \(I_H\) and \(I_R\) (\(I_H\cap I_R=\varnothing\), \(I_H\cup I_R=\{1,\dots,k\}\)). We define
\(z_H(x)\) and \(z_R(x)\) as the subvectors of \(z(x)\) supported on \(I_H\) and \(I_R\), respectively, so that $z(x)$ can be expressed as:
\[
z(x)\;=\;z_H(x)\;\oplus\;z_R(x).
\]
This is a coordinate-disjoint split in the PCA coordinate system. We then train the residual calibrator also as a linear function on the residual features as,
\[
r\big(z_R(x)\big) \;=\; w_r^\top z_R(x) + b_r,
\]
where \(w_r\in\mathbb{R}^{|I_R|}\) and \(b_r\in\mathbb{R}\). The base predictor \(h\) is a linear logit model on \(z_H(x)\), and can be formulated as
\[
h\big(z_H(x)\big) \;=\; \sigma\!\left(w_h^\top z_H(x) + b_h\right).
\]
With \(h\) frozen, we train \(r\) by minimizing the loss in \Cref{residualpred_objective}, which fits the remaining error not captured by \(h\). The final prediction is then formulated as:
\[
\hat{y}(x)
\;=\; \sigma\!\Big(
\underbrace{w_h^\top z_H(x) + b_h}_{\text{base logit}}
\;+
\underbrace{w_r^\top z_R(x) + b_r}_{\text{residual logit}}
\Big).
\]
Because \(z_H\) and \(z_R\) use disjoint PCA dimensions, the contributions from \(h\) and \(r\) are additively separable at the logit level, and each coefficient in \(w_h\) and \(w_r\) can be interpreted as the marginal effect of its corresponding feature. 

Residual calibration is motivated by a pragmatic mismatch in molecular property prediction: a linear predictor \(h\) offers explainable feature attributions but can underfit data, leaving predictive error in the remaining representation. We therefore train \(r\) on the complementary feature block and add its output in logit space. This correction refines the decision boundary without contaminating the main effects made by \(h\). This mechanism typically improves predictive accuracy because \(r\) explicitly predicts the residual error not explained by \(h\), while the base predictor is kept fixed to prevent erroneous reallocation of feature contributions. Beyond improving accuracy, residual calibration enhances explainability by recovering missing or mispredicted functional groups, iteratively aligning predictions with chemically consistent substructures. We formalize this guarantee in the following remark (demonstration provided in the Supplementary Material B):

\begin{remark}
Let \( \mathcal{X} \) and \( \mathcal{Y} \) be the input and output spaces. Let \( f: \mathcal{X} \rightarrow \mathbb{R}^{d} \) be a pre-trained feature mapping, and let \(z(x)\in\mathbb{R}^k\) be its PCA representation. Let \( h: \mathbb{R}^{|I_H|} \rightarrow \mathcal{Y} \) be a linear mapping operating on the explainable PCA coordinates \( z_H(x) \). The residual calibrator \( r: \mathbb{R}^{|I_R|} \rightarrow \mathcal{Y} \), defined on the residual PCA coordinates \( z_R(x) \), captures the remaining variance not explained by \( h \) in an interpretable way, thereby preserving the overall model's explainability.
\end{remark} \label{theorem2}

\textbf{Quantifiable Functional Group Contribution.}
We tokenize Group SELFIES at the functional‑group level and encode each group \(x_j\) with the LLM individually to obtain semantically meaningful embeddings. We then aggregate these group‑wise embeddings and train a linear predictor, where the coefficient \(w_j\) associated with \(x_j\) quantifies its impact on the target property \(y\). Due to the linearity, \(w_j\,[f_H(\tilde{x})]_j\) measures per‑instance contribution score of the \(j\)-th functional group to the predicted property, providing a meaningful attribution under the linear modeling assumptions~\cite{rao1995linear}. Furthermore, this enables precise identification of contributory substructures and their interactions from high‑dimensional LLM representations. Technically, we map 1D text‑based molecules back to their 2D molecular graphs using the contribution scores. By attributing significant functional groups, we identify important nodes and edges in the molecular graph and connect them to highlight contributory substructures. Based on this, we can assess whether the revealed substructure matches the ground truth.

\section{Experiments}\label{sec:experiments}

\subsection{Experimental Setup} \label{experimental settings}

\begin{table*}[t]
    \small
    \centering
    \caption{Classification accuracy over seven datasets (\%). The best and second-best results are shown in bold and underlined, respectively.}
    \vspace{-0.6em}  
    \label{tab:accuracy}
    \renewcommand{\arraystretch}{0.7} 
    \setlength{\tabcolsep}{6pt} 
    \resizebox{\textwidth}{!}{ 
    \begin{tabular}{cccccccc}
        \toprule       
        Methods & Mutag & Mutagen & PTC-FM & PTC-FR & PTC-MM & PTC-MR & Liver \\ \midrule
        \scriptsize MolCLR \cite{moiclr} & \scriptsize 85.7{\tiny ± 0.4} & \scriptsize 79.2{\tiny ± 0.6} & \scriptsize 59.7{\tiny ± 0.5} & \scriptsize 62.8{\tiny ± 0.5} & \scriptsize 62.4{\tiny ± 0.4} & \scriptsize 57.6{\tiny ± 0.5} & \scriptsize 45.2{\tiny ± 0.9} \\
        \scriptsize CGIB \cite{cgib} & \scriptsize 87.1{\tiny ± 0.2} & \scriptsize 81.5{\tiny ± 0.3} & \scriptsize 62.3{\tiny ± 0.5} & \scriptsize 65.7{\tiny ± 0.5} & \scriptsize 65.1{\tiny ± 0.5} & \scriptsize 61.1{\tiny ± 0.7} & \scriptsize 46.9{\tiny ± 0.7} \\
        \scriptsize MoSE \cite{mose} & \scriptsize 88.9{\tiny ± 0.4} & \underline{\scriptsize 83.3{\tiny ± 0.2}} & \underline{\scriptsize 63.3{\tiny ± 0.7}} & \scriptsize 68.0{\tiny ± 0.4} & \scriptsize 67.1{\tiny ± 0.2} & \scriptsize 62.3{\tiny ± 0.6} & \underline{\scriptsize 50.1{\tiny ± 0.6}} \\
        \scriptsize FragNet \cite{fragnet} & \scriptsize 88.5{\tiny ± 0.5} & \scriptsize 82.5{\tiny ± 0.4} & \scriptsize 60.9{\tiny ± 0.8} & \scriptsize 65.7{\tiny ± 0.8} & \underline{\scriptsize 70.3{\tiny ± 0.3}} & \scriptsize \underline{65.0{\tiny ± 0.2}} & \scriptsize 49.6{\tiny ± 0.6} \\
        \scriptsize MomentGNN \cite{momentgnn} & \scriptsize 89.7{\tiny ± 0.4} & \scriptsize 81.5{\tiny ± 0.3} & \scriptsize 59.5{\tiny ± 0.9} & \scriptsize 68.8{\tiny ± 0.7} & \scriptsize 65.4{\tiny ± 0.4} & \scriptsize 60.7{\tiny ± 0.3} & \scriptsize 47.9{\tiny ± 0.9} \\
        \scriptsize S-CGIB \cite{s-cgib} & \scriptsize \textbf{92.0{\tiny ± 0.3}} & \scriptsize 81.3{\tiny ± 0.5} & \scriptsize 60.8{\tiny ± 0.7} & \scriptsize 65.1{\tiny ± 0.6} & \scriptsize 66.3{\tiny ± 0.8} & \scriptsize 64.5{\tiny ± 0.9} & \scriptsize 48.3{\tiny ± 1.0} \\
        
        \scriptsize GraphMVP \cite{graphmvp} & \scriptsize 86.8{\tiny ± 0.2} & \scriptsize 80.4{\tiny ± 0.4} & \scriptsize 58.6{\tiny ± 0.4} & \scriptsize 64.2{\tiny ± 0.3} & \scriptsize 63.7{\tiny ± 0.5} & \scriptsize 62.0{\tiny ± 0.4} & \scriptsize 46.9{\tiny ± 0.4} \\ \midrule
        \scriptsize LLAMA3.1-8b \cite{dubey2024llama} & \scriptsize 67.6{\tiny ± 3.4} & \scriptsize 50.7{\tiny ± 3.6} & \scriptsize 49.6{\tiny ± 2.6} & \scriptsize 46.2{\tiny ± 3.8} & \scriptsize 42.0{\tiny ± 2.8} & \scriptsize 47.5{\tiny ± 2.8} & \scriptsize 42.2{\tiny ± 2.2} \\
        \scriptsize GPT-4o \cite{achiam2023gpt} & \scriptsize 73.5{\tiny ± 3.6} & \scriptsize 51.2{\tiny ± 0.5} & \scriptsize 52.7{\tiny ± 2.3} & \scriptsize 53.8{\tiny ± 2.9} & \scriptsize 48.8{\tiny ± 2.4} & \scriptsize 53.7{\tiny ± 1.8} & \scriptsize 44.5{\tiny ± 2.5} \\
        \scriptsize ChemBERTa-2 \cite{ahmad2022chemberta} & \scriptsize 87.3{\tiny ± 2.7} & \scriptsize 77.6{\tiny ± 2.2} & \scriptsize 59.2{\tiny ± 1.9} & \scriptsize 64.8{\tiny ± 2.2} & \scriptsize 59.7{\tiny ± 2.8} & \scriptsize 59.8{\tiny ± 2.4} & \scriptsize 46.3{\tiny ± 2.3} \\ \midrule

        \scriptsize Logistic Regression & \scriptsize 58.3{\tiny ± 1.2} & \scriptsize 55.4{\tiny ± 0.8} & \scriptsize 48.4{\tiny ± 1.1} & \scriptsize 48.3{\tiny ± 1.0} & \scriptsize 48.7{\tiny ± 1.1} & \scriptsize 44.9{\tiny ± 1.0} & \scriptsize 32.5{\tiny ± 0.5} \\
        \scriptsize Decision Tree \cite{quinlan1986induction} & \scriptsize 60.8{\tiny ± 1.7} & \scriptsize 58.6{\tiny ± 1.5} & \scriptsize 43.3{\tiny ± 1.0} & \scriptsize 46.1{\tiny ± 0.7} & \scriptsize 47.2{\tiny ± 0.7} & \scriptsize 43.5{\tiny ± 0.5} & \scriptsize 36.9{\tiny ± 0.8} \\
        \scriptsize Random Forest \cite{breiman2001random} & \scriptsize 64.6{\tiny ± 1.9} & \scriptsize 60.6{\tiny ± 1.5} & \scriptsize 46.9{\tiny ± 1.2} & \scriptsize 51.4{\tiny ± 1.5} & \scriptsize 51.3{\tiny ± 1.8} & \scriptsize 46.4{\tiny ± 1.1} & \scriptsize 34.8{\tiny ± 1.9} \\
        \scriptsize XGBoost \cite{chen2016xgboost} & \scriptsize 66.9{\tiny ± 1.2} & \scriptsize 67.6{\tiny ± 1.4} & \scriptsize 51.4{\tiny ± 1.3} & \scriptsize 53.1{\tiny ± 1.4} & \scriptsize 55.8{\tiny ± 1.2} & \scriptsize 49.3{\tiny ± 2.1} & \scriptsize 38.5{\tiny ± 1.8} \\ \midrule

        \rowcolor{cyan!6}
        \scriptsize \textit{w/o} Calibration & \scriptsize 86.1{\tiny ± 2.2} & \scriptsize 74.4{\tiny ± 1.0} & \scriptsize 59.7{\tiny ± 2.1} & \underline{\scriptsize 68.9{\tiny ± 1.9}} & \scriptsize 69.3{\tiny ± 2.7} & \scriptsize 61.2{\tiny ± 2.4} & \scriptsize 45.0{\tiny ± 2.0} \\
        \rowcolor{cyan!6}
        \scriptsize {\bf \textit{w/} Calibration (Ours)} & \underline{\scriptsize 91.6{\tiny ± 2.0}} & \textbf{\scriptsize 83.7{\tiny ± 0.9}} & \textbf{\scriptsize 64.2{\tiny ± 1.4}} & \textbf{\scriptsize 74.4{\tiny ± 1.9}} & \textbf{\scriptsize 76.4{\tiny ± 1.8}} & \textbf{\scriptsize 68.4{\tiny ± 2.3}} & \textbf{\scriptsize 54.9{\tiny ± 2.4}} \\
        \bottomrule
    \end{tabular}
    }
    \vspace{-3pt}
\end{table*}

\begin{table*}[t]
    \small
    \centering
    \caption{Explanation accuracy over seven datasets (\%). The \textbf{best} and \underline{second-best} results are marked in bold and underlined, respectively. As LLMs produce uncalibrated, non-discriminative probabilities unsuitable for AUC-based ranking, we omit their explanation accuracy from this table.}
    \vspace{-0.6em}  
    \label{tab:explanation_accuracy}
    \renewcommand{\arraystretch}{0.7} 
    \setlength{\tabcolsep}{6pt} 
    \resizebox{\textwidth}{!}{ 
    \begin{tabular}{cccccccc}
        \toprule
        Methods & Mutag & Mutagen & PTC-FM & PTC-FR & PTC-MM & PTC-MR & Liver \\ \midrule
        
        \scriptsize MolCLR \cite{moiclr} & \scriptsize 85.1{\tiny ± 0.2} & \scriptsize 76.7{\tiny ± 0.3} & \scriptsize 64.3{\tiny ± 0.6} & \scriptsize 63.7{\tiny ± 0.3} & \scriptsize 64.1{\tiny ± 0.2} & \scriptsize 65.9{\tiny ± 0.8} & \scriptsize 67.8{\tiny ± 0.3} \\
        
        \scriptsize CGIB \cite{cgib} & \scriptsize 86.1{\tiny ± 0.3} & \scriptsize 73.6{\tiny ± 0.1} & \scriptsize 62.3{\tiny ± 0.8} & \scriptsize 66.0{\tiny ± 0.6} & \underline{\scriptsize 69.0{\tiny ± 0.6}} & \scriptsize 62.0{\tiny ± 0.4} & \scriptsize 69.1{\tiny ± 0.6} \\
        
        \scriptsize MoSE \cite{mose} & \scriptsize 79.5{\tiny ± 0.5} & \scriptsize 81.5{\tiny ± 0.5} & \underline{\scriptsize 69.0{\tiny ± 0.3}} & \scriptsize 66.2{\tiny ± 0.8} & \scriptsize 68.2{\tiny ± 0.5} & \scriptsize 59.3{\tiny ± 0.5} & \scriptsize 68.9{\tiny ± 0.8} \\
        
        \scriptsize FragNet \cite{fragnet} & \scriptsize 83.2{\tiny ± 0.4} & \scriptsize 80.7{\tiny ± 0.4} & \scriptsize 65.1{\tiny ± 0.2} & \scriptsize 69.9{\tiny ± 0.4} & \scriptsize 65.7{\tiny ± 0.4} & \scriptsize 61.1{\tiny ± 0.7} & \scriptsize 62.4{\tiny ± 0.4} \\
        
        \scriptsize MomentGNN \cite{momentgnn} & \scriptsize 81.0{\tiny ± 0.3} & \scriptsize 78.2{\tiny ± 0.3} & \scriptsize 67.5{\tiny ± 0.2} & \scriptsize 62.5{\tiny ± 0.6} & \scriptsize 63.1{\tiny ± 0.2} & \underline{\scriptsize 70.7{\tiny ± 0.5}} & \scriptsize 70.7{\tiny ± 0.5} \\
        
        \scriptsize S-CGIB \cite{s-cgib} & \scriptsize 86.7{\tiny ± 0.4} & \scriptsize 77.3{\tiny ± 0.1} & \scriptsize 62.4{\tiny ± 0.9} & \scriptsize 64.9{\tiny ± 0.9} & \scriptsize 65.5{\tiny ± 0.8} & \scriptsize 57.5{\tiny ± 0.3} & \scriptsize 64.2{\tiny ± 0.9} \\
        
        \scriptsize GraphMVP \cite{graphmvp} & \scriptsize 82.4{\tiny ± 0.2} & \underline{\scriptsize 82.2{\tiny ± 0.6}} & \scriptsize 61.6{\tiny ± 0.4} & \underline{\scriptsize 71.0{\tiny ± 0.3}} & \scriptsize 63.4{\tiny ± 0.5} & \scriptsize 68.3{\tiny ± 0.3} & \scriptsize 66.3{\tiny ± 0.3} \\ \midrule
        
        \scriptsize Logistic Regression & \scriptsize 59.2{\tiny ± 0.4} & \scriptsize 50.6{\tiny ± 0.9} & \scriptsize 54.4{\tiny ± 0.3} & \scriptsize 47.7{\tiny ± 0.8} & \scriptsize 49.9{\tiny ± 0.7} & \scriptsize 44.3{\tiny ± 0.7} & \scriptsize 53.8{\tiny ± 0.7} \\
        \scriptsize Decision Tree \cite{quinlan1986induction} & \scriptsize 61.2{\tiny ± 0.2} & \scriptsize 55.7{\tiny ± 1.0} & \scriptsize 56.7{\tiny ± 0.8} & \scriptsize 46.4{\tiny ± 1.1} & \scriptsize 48.1{\tiny ± 0.9} & \scriptsize 39.9{\tiny ± 0.8} & \scriptsize 56.4{\tiny ± 1.0} \\
        \scriptsize Random Forest \cite{breiman2001random} & \scriptsize 66.7{\tiny ± 1.2} & \scriptsize 57.2{\tiny ± 1.2} & \scriptsize 59.9{\tiny ± 1.7} & \scriptsize 50.9{\tiny ± 1.2} & \scriptsize 55.0{\tiny ± 0.8} & \scriptsize 46.6{\tiny ± 1.1} & \scriptsize 60.7{\tiny ± 1.4} \\
        \scriptsize XGBoost \cite{chen2016xgboost} & \scriptsize 65.2{\tiny ± 1.2} & \scriptsize 61.3{\tiny ± 1.1} & \scriptsize 58.5{\tiny ± 1.8} & \scriptsize 49.4{\tiny ± 1.8} & \scriptsize 51.6{\tiny ± 1.3} & \scriptsize 50.2{\tiny ± 0.8} & \scriptsize 69.0{\tiny ± 1.4} \\ \midrule
        \rowcolor{cyan!6}
        \scriptsize \textit{w/o} Calibration & \underline{\scriptsize 90.0{\tiny ± 0.9}} & \scriptsize 77.7{\tiny ± 1.0} & \scriptsize 68.0{\tiny ± 1.7} & \scriptsize 66.6{\tiny ± 1.1} & \scriptsize 62.0{\tiny ± 1.5} & \scriptsize 67.5{\tiny ± 1.5} & \underline{\scriptsize 72.0{\tiny ± 2.0}} \\
        \rowcolor{cyan!6}
        \scriptsize {\bf \textit{w/} Calibration (Ours)} & \textbf{\scriptsize 92.6{\tiny ± 1.7}} & \textbf{\scriptsize 89.0{\tiny ± 1.2}} & \textbf{\scriptsize 77.9{\tiny ± 1.5}} & \textbf{\scriptsize 79.3{\tiny ± 1.4}} & \textbf{\scriptsize 72.3{\tiny ± 1.7}} & \textbf{\scriptsize 73.4{\tiny ± 1.3}} & \textbf{\scriptsize 80.3{\tiny ± 1.4}} \\
        \bottomrule
    \end{tabular}
    }
\end{table*}

\textbf{Datasets.} We evaluate {\em MoleX} on six mutagenicity datasets and one hepatotoxicity dataset. The mutagenicity datasets include Mutag~\cite{debnath1991structure}, Mutagen~\cite{morris2020tudataset}, and the PTC family~\cite{toivonen2003statistical}; the hepatotoxicity dataset is Liver~\cite{liu2015data}. To demonstrate {\em MoleX}'s ability to explain molecular properties using chemically meaningful substructures, we introduce ground truth, substructures verified by domain experts to significantly impact molecular properties. Ground truth annotations for mutagenicity are from~\cite{lin2022orphicx, debnath1991structure}, and for hepatotoxicity from~\cite{cheng2023group}. To demonstrate the effectiveness of \textit{MoleX}, we further evaluate it on six MoleculeNet datasets: BBBP, Tox21, SIDER, ClinTox, BACE, and HIV~\cite{wu2018moleculenet}. MoleculeNet includes tasks influenced by nonlinear interactions, stereochemistry, and scaffold bias, making it a relatively challenging benchmark. Dataset details are provided in the Supplementary Material C.

\textbf{Evaluation Metrics.} In this study, we evaluate the predictive performance, explainability performance, and computational efficiency of {\em MoleX}. In particular, we apply a specific metric to assess each of these aspect. For predictive performance, we define  \(\frac{1}{n} \sum_{i=1}^{n} \mathbb{I}(y^{(i)} = \hat{y}^{(i)})\) to compute the classification accuracy. For explainability performance, we follow GNNExplainer \cite{ying2019gnnexplainer}, treating explanations as binary edge classification and using AUC to measure the accuracy. For computational efficiency, we evaluate the execution time for each method.

Moreover, we also assess explanation accuracy against expert-annotated ground truth. As verified by \cite{lin2022orphicx, debnath1991structure}, the ground truth substructures for six mutagenicity datasets consist of an aromatic group, such as a benzene ring, bonded with another functional group, such as methoxy, oxhydryl, nitro, or carboxyl groups (note that ground truth exists only for the mutagenic class). For the Liver dataset, the chemist‑annotated ground truth includes features such as fused tricyclic saturated hydrocarbon moieties, hydrazines, arylacetic acids, and sulfonamide moieties~\cite{liu2015data}. We compare whether the substructures revealed by MoleX match the ground truth.

\textbf{Baselines.} To extensively compare \emph{MoleX} with advanced methods, we utilize various baselines: (1) GNN baselines (all designed specifically for molecular property prediction), including MolCLR \cite{moiclr}, CGIB \cite{cgib},  MoSE \cite{mose}, FragNet \cite{fragnet}, MomentGNN \cite{momentgnn},  S-CGIB \cite{s-cgib}, and GraphMVP \cite{graphmvp}; (2) LLM baselines, including Llama 3.1-8b \cite{dubey2024llama}, GPT-4o \cite{achiam2023gpt}, and ChemBERTa-2 \cite{ahmad2022chemberta}; (3) explainable machine learning baselines, including logistic regression, decision tree \cite{quinlan1986induction}, XGBoost \cite{chen2016xgboost}, and random forest \cite{breiman2001random}.

\begin{figure*}[t]
\setlength{\abovecaptionskip}{5pt} 
\setlength{\belowcaptionskip}{5pt} 
\centering
\includegraphics[width=0.8\textwidth]{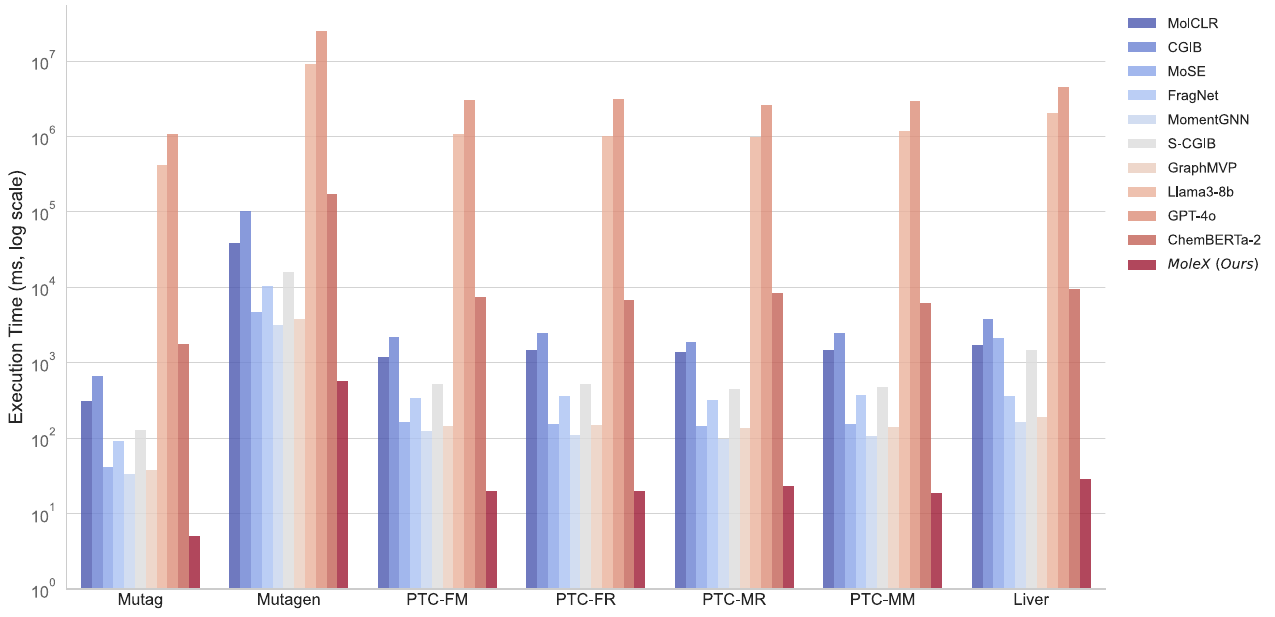} 
\caption{Execution time comparison on seven datasets. \textit{MoleX} achieves the best efficiency.}
\label{fig:efficiency}
\vspace{-5mm}
\end{figure*}

\textbf{Implementations.} Our model continues pre‑training on the full ZINC dataset~\cite{irwin2012zinc} using ChemBERTa‑2, with $15\%$ of tokens in each input randomly masked. We then fine-tune it on Mutag, Mutagen, PTC-FM, PTC-FR, PTC-MM, PTC-MR, and Liver (in Group SELFIES). To evaluate the performance, we compute the average and standard deviation of each metric for each method after 20 rounds of execution. The details of parameter settings is offered in the Supplementary Material D.

\subsection{Results and Analysis} \label{experimental results}
\textbf{Predictive Performance.} \Cref{tab:accuracy} presents a comparison of predictive performance across different methods. {\em MoleX} outperforms all baselines, demonstrating strong robustness and generalizability. By leveraging linear predictor to learn decomposed LLM representations, it achieves $16.9\%$ and $23.1\%$ higher average accuracy compared to LLM and explainable method baselines, respectively. These results highlight the effectiveness of learning with explainable, chemically consistent concepts. Incorporating residual calibration further boosts the average classification accuracy by $11.2\%$ across all datasets. On Mutag, the classification accuracy of the linear predictor improves by $27.8\%$ after integrating LLM-derived representations, followed by an additional $5.5\%$ gain with residual calibration. These results demonstrate that maximizing task-relevant semantics and applying residual calibration enables even simple linear predictor to outperform both GNNs and LLMs. More predictive performance results on six MoleculeNet datasets~\cite{wu2018moleculenet} are provided in the Supplementary Material J.

\begin{table}[t]
  \centering
  \caption{Classification and explanation accuracy (\%) \emph{without} PCA over seven datasets.}
  \small
  \setlength{\tabcolsep}{8pt} 
  \renewcommand{\arraystretch}{1.0} 
  \begin{tabular}{ccc}
    \toprule
    Dataset & Classification Accuracy & Explanation Accuracy \\
    \midrule
    Mutag   & 94.9 {\footnotesize $\pm$ 1.6} & 96.1 {\footnotesize $\pm$ 3.0} \\
    Mutagen & 86.4 {\footnotesize $\pm$ 1.4} & 91.2 {\footnotesize $\pm$ 1.6} \\
    PTC-FR  & 78.7 {\footnotesize $\pm$ 1.2} & 82.7 {\footnotesize $\pm$ 1.7} \\
    PTC-FM  & 68.1 {\footnotesize $\pm$ 1.5} & 81.1 {\footnotesize $\pm$ 2.0} \\
    PTC-MR  & 70.5 {\footnotesize $\pm$ 1.7} & 76.5 {\footnotesize $\pm$ 2.6} \\
    PTC-MM  & 80.9 {\footnotesize $\pm$ 2.7} & 75.3 {\footnotesize $\pm$ 2.2} \\
    Liver   & 57.3 {\footnotesize $\pm$ 1.6} & 83.8 {\footnotesize $\pm$ 1.9} \\
    \bottomrule
  \end{tabular}
  \label{tab:pca-works-table}
  \vspace{-4mm}
\end{table}

\textbf{Explainability Performance.} \Cref{tab:explanation_accuracy} reports the explanation accuracy across different methods. Through learning decomposed LLM representations, {\em MoleX} significantly outperforms all baselines in explainability. Residual calibration further improves performance, yielding an $8.8\%$ average gain by iteratively correcting mispredicted functional groups and reinforcing chemically accurate ones. On Mutag, the explanation accuracy of the linear predictor improves by $33.4\%$ with LLM decomposition and residual calibration. While other methods perform well on simpler datasets like Mutag but fail on more complex ones, {\em MoleX} achieves $13.2\%$ higher classification and $16.9\%$ higher explanation accuracy on Liver, demonstrating its ability to \textit{capture the complexity of molecular representations} even with a linear mapping. More explainability performance results on six MoleculeNet datasets~\cite{wu2018moleculenet} are provided in the Supplementary Material J.

\Cref{fig:umap} visualizes the explanation for a molecule from the Mutag. Domain experts attribute its mutagenicity to an aromatic ring (\textit{e.g.,} benzene) bonded to a functional group such as nitro or carbonyl. {\em MoleX} accurately reveals this chemically consistent substructure by learning a linear mapping over decomposed LLM representations that encode high-level chemical concepts. Unlike baseline methods that highlight isolated atoms or bonds without structural coherence---for example, PGExplainer marks scattered atoms across multiple benzene rings---{\em MoleX} yields explanations that are chemically meaningful and structurally complete. More importantly, the uncalibrated linear predictor includes extraneous atoms, while residual calibration corrects these mispredictions and aligns the output with ground truth. This suggests the critical role of calibration in refining explanations and enhancing the explainability of LLM-derived features. Contribution scores further indicate functional group interactions, with the benzene–nitro substructure (\textit{i.e.,} consistent with the ground truth) receiving the highest score, showing its importance in mutagenicity. More visualizations are offered in the Supplementary Material G.

\begin{figure*}[t]
  \centering
  \captionsetup{aboveskip=4pt} 

  \begin{subfigure}{0.9\textwidth}
    \centering
    \includegraphics[width=\linewidth]{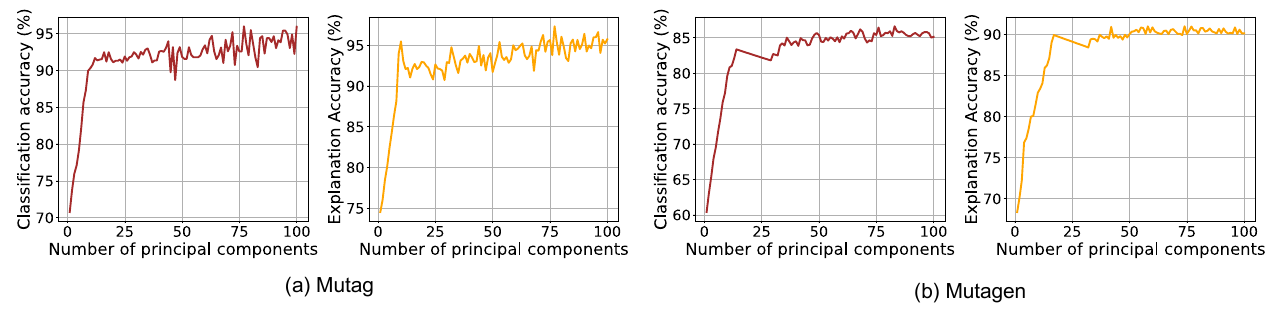}
    \caption{Effect of the number of principal components on model performance.}
    \label{ablation_pca}
  \end{subfigure}

  \vspace{6pt}

  \begin{subfigure}{0.9\textwidth}
    \centering
    \includegraphics[width=\linewidth]{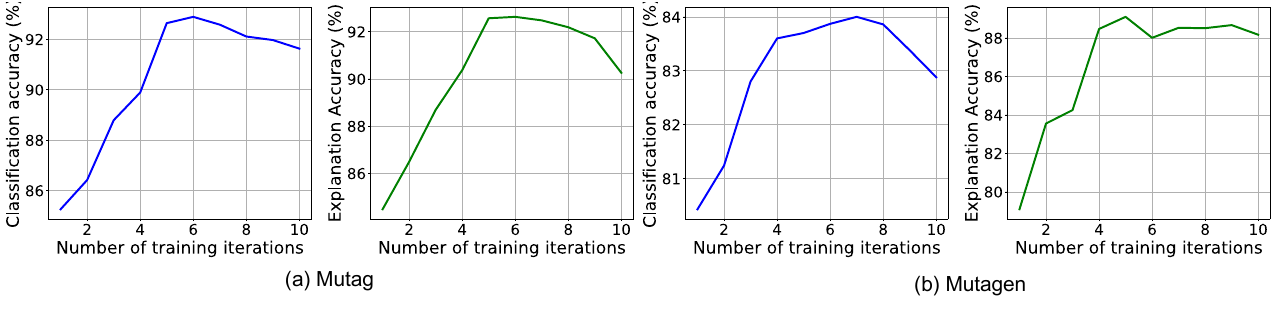}
    \caption{Effect of the number of residual calibration training iterations on model performance.}
    \label{ablation_residual}
  \end{subfigure}

  \vspace{-4mm} 
\end{figure*}

\textbf{Computational Efficiency.} \Cref{fig:efficiency} compares execution time across methods. Unlike approaches requiring iterative network optimization, {\em MoleX} achieves significantly faster inference via a linear predictor over LLM-derived chemical features. It is at least $10\times$ faster than GNNs and over $300\times$ faster than LLM, while delivering superior performance. Across all datasets, {\em MoleX} consistently achieves the lowest inference time, demonstrating feasibility for large-scale molecular analysis. By eliminating iterative parameter updates and avoiding the memory overhead of gradient-based optimization, it also substantially reduces GPU usage. This indicates that a linear mapping over structurally decomposed LLM representations improves efficiency in molecular representation learning.

\subsection{Ablation Studies} \label{ablation}

\textbf{How to choose the optimal number of principal components?}  
To obtain compact yet informative representations, we apply PCA to reduce the dimensionality of LLM-derived features. As shown in \Cref{ablation_pca}, both predictive and explanation accuracy improve with more components but begin to plateau beyond 20, suggesting that additional components contribute minimally to model performance. This observation implies that the dominant molecular semantics are captured within the top 20 principal components, while further dimensions primarily encode noise or low-variance information. Including more components increases model complexity and hampers explainability without significant performance gain. Therefore, we retain the top 20 principal components to capture the most significant variance in molecular properties while maintaining a balance between accuracy and explainability.

\textbf{How does dimensionality reduction contribute to results?}  
We show that PCA effectively preserves the most significant dimensions in LLM‑derived molecular features. For comparison, we also report model performance across seven datasets \textit{without} dimensionality reduction. As shown in \Cref{tab:pca-works-table}, using only the top 20 principal components yields performance within $5\%$ of that achieved using all 384 components. This result suggests that PCA effectively identifies and retains the most task-relevant semantic directions, while filtering out low-variance or noisy components that contribute little to prediction. By projecting high-dimensional representations onto a compact and explainable subspace, PCA not only preserves predictive power but also enhances model transparency and robustness. The near-equivalent performance under substantial dimensionality reduction highlights the redundancy of many raw LLM representation and supports the hypothesis that molecular properties are largely governed by a lower-dimensional semantic linear structure. In light of this, PCA serves as an effective mechanism for improving explainability without sacrificing accuracy.

\begin{figure*}[t]
\centering
\captionsetup{aboveskip=4pt} 
\includegraphics[width=0.96\textwidth]{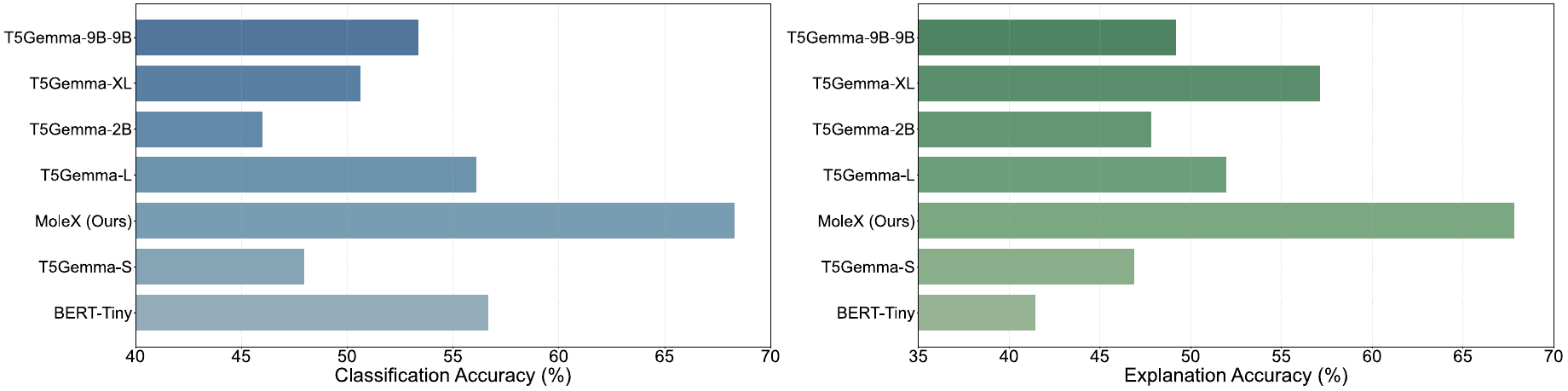} 
\caption{Effect of LLM model size on downstream performance.}
\label{parameter}
\vspace{-2mm} 
\end{figure*}

\begin{table*}[h]
\centering
\captionsetup{width=0.9\textwidth} 
\caption{Classification accuracy across different machine learning models over seven datasets (\%).}
\small
\resizebox{0.9\textwidth}{!}{ 
\begin{tabular}{cccccccc}
\toprule
Method & Mutag & Mutagen & PTC-FR & PTC-FM & PTC-MR & PTC-MM & Liver \\
\midrule
Ridge Regression & 90.7\scriptsize{±1.2} & 84.1\scriptsize{±1.3} & 72.4\scriptsize{±2.0} & 65.2\scriptsize{±2.0} & 69.8\scriptsize{±1.4} & 77.5\scriptsize{±1.5} & 58.1\scriptsize{±1.6} \\
LASSO & 91.9\scriptsize{±1.7} & 84.4\scriptsize{±0.7} & 75.1\scriptsize{±2.1} & 65.8\scriptsize{±1.7} & 65.2\scriptsize{±0.9} & 74.2\scriptsize{±1.2} & 58.7\scriptsize{±1.8} \\
Linear Discriminant Analysis & 89.9\scriptsize{±1.9} & 83.6\scriptsize{±1.2} & 75.2\scriptsize{±1.9} & 65.7\scriptsize{±1.9} & 69.3\scriptsize{±1.8} & 76.8\scriptsize{±2.0} & 57.7\scriptsize{±1.3} \\
Support Vector Machine & 93.9\scriptsize{±1.6} & 86.6\scriptsize{±1.5} & 73.4\scriptsize{±1.9} & 69.3\scriptsize{±2.6} & 69.5\scriptsize{±2.0} & 78.6\scriptsize{±1.3} & 61.5\scriptsize{±2.9} \\
Decision Tree & 89.7\scriptsize{±2.1} & 79.5\scriptsize{±1.2} & 72.4\scriptsize{±1.8} & 64.3\scriptsize{±2.1} & 68.5\scriptsize{±1.5} & 74.4\scriptsize{±1.4} & 59.5\scriptsize{±2.2} \\
Random Forest & 92.8\scriptsize{±2.7} & 84.4\scriptsize{±1.7} & 77.3\scriptsize{±2.1} & 68.6\scriptsize{±2.5} & 71.0\scriptsize{±2.2} & 77.2\scriptsize{±2.1} & 62.7\scriptsize{±2.7} \\
Gradient Boosting Machine & 94.8\scriptsize{±2.1} & 85.3\scriptsize{±1.9} & 78.9\scriptsize{±1.9} & 69.4\scriptsize{±2.8} & 72.2\scriptsize{±2.1} & 79.2\scriptsize{±1.9} & 63.9\scriptsize{±2.6} \\
XGBoost & 94.6\scriptsize{±2.3} & 85.0\scriptsize{±2.0} & 78.7\scriptsize{±2.2} & 70.1\scriptsize{±2.3} & 73.4\scriptsize{±2.9} & 78.1\scriptsize{±2.1} & 63.0\scriptsize{±2.3} \\
{\em MoleX} (Ours) & 91.6\scriptsize{±2.0} & 83.7\scriptsize{±0.9} & 74.4\scriptsize{±1.9} & 64.2\scriptsize{±1.4} & 68.4\scriptsize{±2.3} & 76.4\scriptsize{±1.8} & 54.9\scriptsize{±2.4} \\
\bottomrule
\end{tabular}
}
\label{tab:other ml models acc}
\end{table*}

\begin{table*}[h]
\centering
\captionsetup{width=0.9\textwidth} 
\caption{Explanation accuracy across different machine learning models over seven datasets (\%).}
\small
\resizebox{0.9\textwidth}{!}{ 
\begin{tabular}{cccccccc}
\toprule
Method & Mutag & Mutagen & PTC-FR & PTC-FM & PTC-MR & PTC-MM & Liver \\
\midrule
Ridge Regression & 92.8\scriptsize{±1.1} & 89.5\scriptsize{±1.3} & 79.0\scriptsize{±1.2} & 78.1\scriptsize{±1.6} & 72.5\scriptsize{±2.5} & 69.7\scriptsize{±2.3} & 82.4\scriptsize{±1.7} \\
LASSO & 92.3\scriptsize{±1.5} & 89.6\scriptsize{±0.9} & 76.9\scriptsize{±1.8} & 81.2\scriptsize{±1.9} & 70.4\scriptsize{±2.3} & 70.7\scriptsize{±2.1} & 81.3\scriptsize{±1.8} \\
Linear Discriminant Analysis & 92.9\scriptsize{±1.8} & 88.5\scriptsize{±1.9} & 80.7\scriptsize{±2.3} & 80.1\scriptsize{±2.2} & 71.7\scriptsize{±2.8} & 71.3\scriptsize{±1.6} & 87.8\scriptsize{±1.6} \\
Support Vector Machine & 92.0\scriptsize{±1.7} & 92.0\scriptsize{±1.6} & 84.7\scriptsize{±2.2} & 86.3\scriptsize{±2.0} & 80.1\scriptsize{±2.3} & 76.0\scriptsize{±2.3} & 81.9\scriptsize{±2.1} \\
Decision Tree & 87.6\scriptsize{±1.9} & 89.1\scriptsize{±1.5} & 78.6\scriptsize{±2.0} & 80.7\scriptsize{±1.6} & 73.1\scriptsize{±2.1} & 74.2\scriptsize{±1.8} & 76.0\scriptsize{±1.8} \\
Random Forest & 93.2\scriptsize{±1.9} & 90.5\scriptsize{±1.8} & 82.1\scriptsize{±2.1} & 84.2\scriptsize{±2.2} & 74.2\scriptsize{±2.0} & 74.5\scriptsize{±2.1} & 81.2\scriptsize{±2.0} \\
Gradient Boosting Machine & 92.7\scriptsize{±2.2} & 92.4\scriptsize{±1.5} & 82.9\scriptsize{±2.3} & 85.2\scriptsize{±2.4} & 73.9\scriptsize{±2.9} & 77.7\scriptsize{±2.6} & 84.5\scriptsize{±2.4} \\
XGBoost & 95.6\scriptsize{±1.8} & 90.7\scriptsize{±1.7} & 84.0\scriptsize{±2.2} & 82.0\scriptsize{±2.3} & 74.4\scriptsize{±2.7} & 77.4\scriptsize{±2.2} & 86.2\scriptsize{±2.5} \\
{\em MoleX} (Ours) & 92.6\scriptsize{±1.7} & 89.0\scriptsize{±0.9} & 79.3\scriptsize{±2.6} & 77.9\scriptsize{±2.6} & 73.4\scriptsize{±2.8} & 72.3\scriptsize{±3.0} & 80.3\scriptsize{±2.5} \\
\bottomrule
\end{tabular}
}
\label{tab:other ml models auc}
\vspace{-3mm} 
\end{table*}

\textbf{Does the Residual Calibrator Improves Model Performance by training with more iterations?}  
We adopt the training objective defined in \Cref{residualpred_objective} to learn a residual calibrator that incrementally corrects the prediction errors made by the base linear predictor. To understand how the number of training iterations affects the effectiveness of the calibrator, we conduct an empirical analysis across four datasets: Mutag, Mutagen, PTC-MR, and Liver. As shown in \Cref{ablation_residual}, model performance consistently improves with additional training iterations, indicating that repeated optimization over our designed loss effectively enhances predictive accuracy. Nevertheless, beyond a certain threshold, the model begins to overfit the training residuals, leading to a decline in generalization performance. This observation suggests a trade-off between correction capacity and overfitting risk. Empirically, we find that $5$ training iterations achieve the best performance across datasets. This aligns with our theoretical analysis (detailed in the Supplementary Material E), which shows that repeated residual updates improve the approximation quality up to a point, beyond which the residual function begins to model noise rather than systematic errors. Thus, careful selection of residual calibration training iterations is crucial to realize its benefits without compromising explainability or generalization.

\begin{figure*}[t]
  \centering
  \captionsetup{aboveskip=4pt} 

  \begin{subfigure}{0.95\textwidth}
    \centering
    \includegraphics[width=\linewidth]{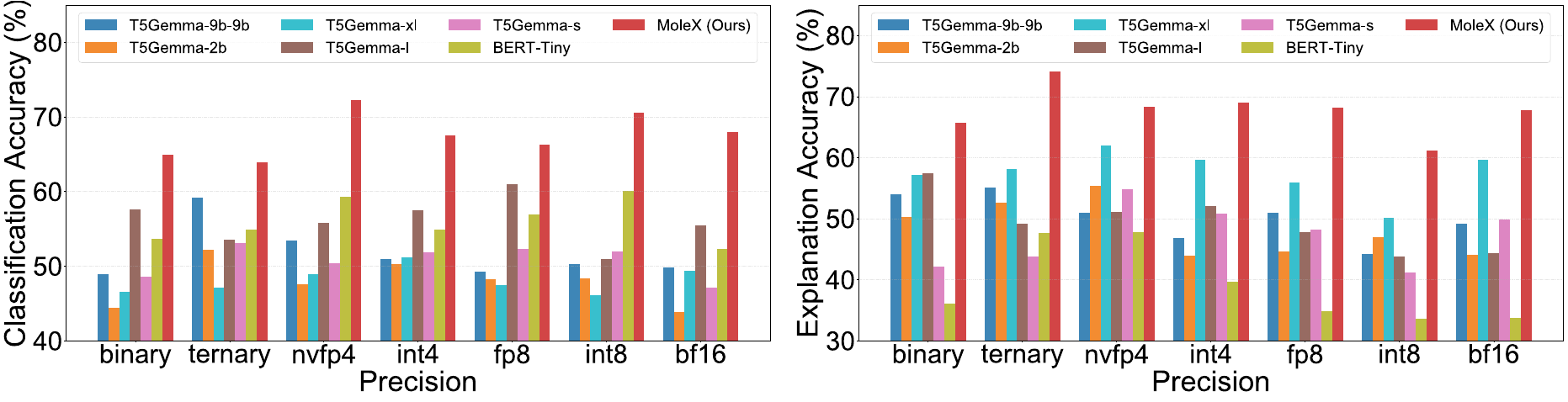}
    \caption{Effect of training precision on downstream performance.}
    \label{precision}
    \vspace{-2mm} 
  \end{subfigure}

  \vspace{6pt} 

  \begin{subfigure}{0.95\textwidth}
    \centering
    \includegraphics[width=\linewidth]{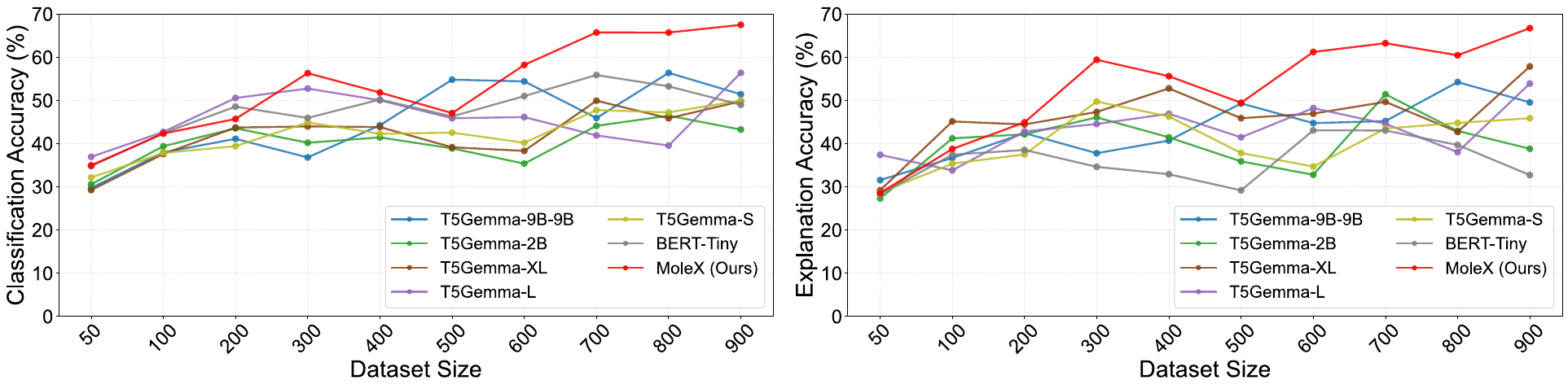}
    \caption{Effect of dataset size on downstream performance.}
    \label{dataset}
  \end{subfigure}

  \vspace{-5mm} 
\end{figure*}

\textbf{Can alternative model architectures learn the LLM representation?}    
To assess the versatility of LLM-derived molecular representations, we evaluate a broad spectrum of machine learning predictors across seven datasets, using both classification and explanation accuracy as evaluation metrics (\Cref{tab:other ml models acc} and \Cref{tab:other ml models auc}). Among shallow linear predictors, ridge regression, LASSO, and LDA perform on par with \textit{MoleX}, indicating that LLM representations linearly encode salient chemical features for property prediction. Simple linear predictors thus suffice to extract task‑relevant signals. With more expressive predictors: decision trees, random forests, and gradient boosting (\textit{e.g.,} XGBoost), we observe consistent gains across all datasets. These models capture more complex  interactions among LLM features, indicating that the representations encode rich chemical semantics beyond what linear predictors can exploit. This supports the view that LLMs implicitly learn transferable domain knowledge that downstream models of varying complexity can harness. 

However, this increased predictive power comes with a trade-off: reduced model explainability. While ensemble models achieve the highest classification accuracy, their black-box nature limits their usefulness in scenarios where explainability is essential, such as scientific discovery or regulatory decision-making. In contrast, linear predictors provide more faithful explanations, particularly when paired with feature attribution (\textit{e.g.,} we can thus clearly attribute molecular properties to specific functional groups). Linearity within LLM latent representations supports clear attribution and downstream performance. It offers a favorable balance between accuracy and explainability, taking advantage of the LLM’s explanatory power without added architectural complexity. Therefore, this permits a principled attribution of how well the LLM encodes molecular semantics for chemically grounded modeling.

Under identical conditions, more complex predictors (\textit{e.g.,} more complicated network architectures or modeling assumptions) may deliver higher predictive accuracy, but at the cost of explainability~\cite{poole1971assumptions, schmidt2018linear}. We thus seek a balance: linearity is a minimal modeling assumption, and when it achieves strong performance, Occam’s razor favors it over added complexity that yields only marginal gains~\cite{murphy2012machine, hastie2015statistical}. This is our motivation to probe whether LLM molecular representations exhibit linear relationships, enabling accurate prediction without sacrificing chemically consistent explanations. The gains in explainability and transparent feature attribution afforded by representational linearity outweigh the small incremental improvements delivered by complex predictors that only modestly surpass linear baselines. Decomposing LLM representations via a concept‑aligned linear projection produces highly explainable yet strongly predictive molecular semantics. Therefore, investigating and characterizing this latent linearity is central to obtaining explainable molecular representation learning.

\subsection{Scaling the Linear Predictor} \label{scaling}
In this section, we evaluate the linear predictor’s predictive and explanation performance under varying configurations. Specifically, we vary the LLM parameter size, training precision, and dataset size. Motivated by neural scaling laws~\cite{kaplan2020scaling}, we examine how scaling training configurations affects the performance of linear predictor on the four PTC family datasets (PTC‑FR, PTC‑FM, PTC‑MR, and PTC‑MM)~\cite{toivonen2003statistical}.

\textbf{Do larger models always deliver better performance?}
Models in \Cref{parameter} are ordered top to bottom by decreasing parameter size. Larger foundation models generally improve both classification and explanation accuracy, though not monotonically. For example, the 9B model, T5‑Gemma‑9B, shows lower classification accuracy than T5‑Gemma‑L and BERT-Tiny, probably because its capacity is not well aligned to molecular structure given limited chemical training data. Smaller models near the bottom achieve reasonable classification but weaker explanation accuracy, indicating limited concept decomposition and higher representation noise. Overall, capacity provides headroom but does not directly improve performance. Task‑adapted training or fine‑tuning is therefore crucial to convert that capacity into better model performance. Essentially, the key to a strong linear predictor is augmenting the foundation LLM with downstream task capability (\textit{e.g.,} via fine‑tuning or post‑training). Merely increasing model size without appropriate data quality, architectural design, or training strategy yields limited gains.

\textbf{Does higher training precision improve performance?}
\Cref{precision} reports downstream performance under different training precisions for the linear predictor. Empirically, higher precision enhances model performance~\cite{kaplan2020scaling}. In general, the int8 and bf16 deliver the most stable gains, whereas low precisions (\textit{e.g.,} binary, ternary) degrade performance due to quantization noise. Mixed formats such as nvfp4 and int4 perform surprisingly close to fp8, suggesting that moderate quantization preserves the dominant information in molecular representations that is crucial for property prediction. \textit{MoleX} is the most robust across precisions, remaining reliable even after quantization. Overall, precision has an irregular impact on downstream performance, with some run‑to‑run variability. Additionally, we observe that larger models are more robust to decreasing precision, whereas smaller models are affected more significantly by precision changes.

\textbf{Does more training data improve performance?}
\Cref{dataset} shows downstream performance for varying training data sizes. In general, adding more training data improves the downstream performance for all models. Increasing the data from $50$ to roughly $300$–$500$ yields the largest gains for both metrics, marking a clear data‑scarce to data‑sufficient transition. Beyond $500$ samples, classification continues to improve but at a slower and model‑specific pace, whereas explanation shows sharper gains. Among them, \textit{MoleX} scales best with data, suggesting that its representations convert added data into more linearly decodable information tied to the molecular property prediction. Overall, more data primarily boosts explanation faithfulness and narrows gaps across models, while classification gains are steadier and show diminishing returns beyond a threshold.

\section{Conclusion and Discussion}\label{sec:conclusion}

This work investigates and leverages latent linearity in LLM‑derived hidden representations to advance molecular property prediction. In particular, we introduce a chemically consistent decomposition that projects molecular embeddings into a concept space spanned by functional‑group vectors, revealing and exploiting hidden linearity in LLMs’ representational geometry. We also propose a residual calibration strategy that captures complex interactions missed by the linear predictor, yielding more accurate predictions and more faithful explanations. Empirically, \textit{MoleX} delivers improved accuracy, explainability, and computational efficiency across multiple benchmark datasets. These results highlight \textit{MoleX} as an effective approach to explainable molecular representation learning and point to a broader path for aligning foundation models with domain science.

However, trade-off between model complexity and performance is still underexplored in molecular representation learning. In this paper, we use a linear predictor, which may underfit complex molecular patterns in some cases. Preliminary comparisons of \textit{MoleX} with more sophisticated models show only marginal gains for the latter. Future work can examine this complexity–performance trade-off in LLM-based molecular representation learning to identify optimal balances between model complexity and downstream performance.

\bibliography{example_paper}

@inproceedings{lin2022orphicx,
  title={Orphicx: A Causality-Inspired Latent Variable Model For Interpreting Graph Neural Networks},
  author={Lin, Wanyu and Lan, Hao and Wang, Hao and Li, Baochun},
  booktitle={Proceedings of the IEEE/CVF Conference on Computer Vision and Pattern Recognition},
  pages={13729--13738},
  year={2022}
}

@article{ying2019gnnexplainer,
  title={Gnnexplainer: Generating Explanations For Graph Neural Networks},
  author={Ying, Zhitao and Bourgeois, Dylan and You, Jiaxuan and Zitnik, Marinka and Leskovec, Jure},
  journal={Advances in neural information processing systems},
  volume={32},
  year={2019}
}

@inproceedings{lin2021generative,
  title={Generative Causal Explanations For Graph Neural Networks},
  author={Lin, Wanyu and Lan, Hao and Li, Baochun},
  booktitle={International Conference on Machine Learning},
  pages={6666--6679},
  year={2021},
  organization={PMLR}
}

@article{wanyu-nips24,
  title={{Explainable Molecular Property Prediction: Aligning Chemical Concepts with Predictions via Language Models}},
  author={Wang, Zhenzhong and Lin, Zehui and Lin, Wanyu and Yang, Ming and Zeng, Minggang and Tan, Kay Chen},
  journal={arXiv preprint arXiv:2405.16041},
  year={2024}
}

@article{chithrananda2020chemberta,
  title={ChemBERTa: Large-Scale Self-Supervised Pretraining for Molecular Property Prediction},
  author={Chithrananda, Seyone and Grand, Gabriel and Ramsundar, Bharath},
  journal={arXiv preprint arXiv:2010.09885},
  year={2020}
}

@article{ross2022large,
  title={{Large-scale chemical language representations capture molecular structure and properties}},
  author={Ross, Jerret and Belgodere, Brian and Chenthamarakshan, Vijil and Padhi, Inkit and Mroueh, Youssef and Das, Payel},
  journal={Nature Machine Intelligence},
  volume={4},
  number={12},
  pages={1256--1264},
  year={2022},
  publisher={Nature Publishing Group UK London}
}

@article{nips24-d4exp,
  title={{D4explainer: In-distribution explanations of graph neural network via discrete denoising diffusion}},
  author={Chen, Jialin and Wu, Shirley and Gupta, Abhijit and Ying, Rex},
  journal={Advances in Neural Information Processing Systems},
  volume={36},
  year={2024}
}

@article{kindermans2019reliability,
  title={The (un) reliability of saliency methods},
  author={Kindermans, Pieter-Jan and Hooker, Sara and Adebayo, Julius and Alber, Maximilian and Sch{\"u}tt, Kristof T and D{\"a}hne, Sven and Erhan, Dumitru and Kim, Been},
  journal={Explainable AI: Interpreting, explaining and visualizing deep learning},
  pages={267--280},
  year={2019},
  publisher={Springer}
}

@article{adebayo2018sanity,
  title={Sanity checks for saliency maps},
  author={Adebayo, Julius and Gilmer, Justin and Muelly, Michael and Goodfellow, Ian and Hardt, Moritz and Kim, Been},
  journal={Advances in neural information processing systems},
  volume={31},
  year={2018}
}

@inproceedings{sundararajan2017axiomatic,
  title={Axiomatic attribution for deep networks},
  author={Sundararajan, Mukund and Taly, Ankur and Yan, Qiqi},
  booktitle={International conference on machine learning},
  pages={3319--3328},
  year={2017},
  organization={PMLR}
}

@inproceedings{ribeiro2016should,
  title={" Why should i trust you?" Explaining the predictions of any classifier},
  author={Ribeiro, Marco Tulio and Singh, Sameer and Guestrin, Carlos},
  booktitle={Proceedings of the 22nd ACM SIGKDD international conference on knowledge discovery and data mining},
  pages={1135--1144},
  year={2016}
}

@inproceedings{jain2019attention,
  title={Attention is not Explanation},
  author={Jain, Sarthak and Wallace, Byron C},
  booktitle={Proceedings of the 2019 Conference of the North American Chapter of the Association for Computational Linguistics: Human Language Technologies, Volume 1 (Long and Short Papers)},
  pages={3543--3556},
  year={2019}
}

@inproceedings{serrano2019attention,
  title={Is Attention Interpretable?},
  author={Serrano, Sofia and Smith, Noah A},
  booktitle={Proceedings of the 57th Annual Meeting of the Association for Computational Linguistics},
  pages={2931--2951},
  year={2019}
}

@inproceedings{hoover2020exbert,
  title={exBERT: A Visual Analysis Tool to Explore Learned Representations in Transformer Models},
  author={Hoover, Benjamin and Strobelt, Hendrik and Gehrmann, Sebastian},
  booktitle={Proceedings of the 58th Annual Meeting of the Association for Computational Linguistics: System Demonstrations},
  pages={187--196},
  year={2020}
}

@inproceedings{agarwal2021towards,
  title={Towards the unification and robustness of perturbation and gradient based explanations},
  author={Agarwal, Sushant and Jabbari, Shahin and Agarwal, Chirag and Upadhyay, Sohini and Wu, Steven and Lakkaraju, Himabindu},
  booktitle={International Conference on Machine Learning},
  pages={110--119},
  year={2021},
  organization={PMLR}
}

@inproceedings{pope2019explainability,
  title={Explainability Methods for Graph Convolutional Neural Networks},
  author={Pope, Phillip E and Kolouri, Soheil and Rostami, Mohammad and Martin, Charles E and Hoffmann, Heiko},
  booktitle={Proceedings of the IEEE/CVF conference on computer vision and pattern recognition},
  pages={10772--10781},
  year={2019}
}

@article{ahmad2022chemberta,
  title={Chemberta-2: Towards chemical foundation models},
  author={Ahmad, Walid and Simon, Elana and Chithrananda, Seyone and Grand, Gabriel and Ramsundar, Bharath},
  journal={arXiv preprint arXiv:2209.01712},
  year={2022}
}

@article{wu2023black,
  title={From black boxes to actionable insights: a perspective on explainable artificial intelligence for scientific discovery},
  author={Wu, Zhenxing and Chen, Jihong and Li, Yitong and Deng, Yafeng and Zhao, Haitao and Hsieh, Chang-Yu and Hou, Tingjun},
  journal={Journal of Chemical Information and Modeling},
  volume={63},
  number={24},
  pages={7617--7627},
  year={2023},
  publisher={ACS Publications}
}

@article{cheng2023group,
  title={Group SELFIES: a robust fragment-based molecular string representation},
  author={Cheng, Austin H and Cai, Andy and Miret, Santiago and Malkomes, Gustavo and Phielipp, Mariano and Aspuru-Guzik, Al{\'a}n},
  journal={Digital Discovery},
  volume={2},
  number={3},
  pages={748--758},
  year={2023},
  publisher={Royal Society of Chemistry}
}

@inproceedings{alemi2022deep,
  title={Deep Variational Information Bottleneck},
  author={Alemi, Alexander A and Fischer, Ian and Dillon, Joshua V and Murphy, Kevin},
  booktitle={International Conference on Learning Representations},
  year={2022}
}

@article{kingma2015variational,
  title={Variational dropout and the local reparameterization trick},
  author={Kingma, Durk P and Salimans, Tim and Welling, Max},
  journal={Advances in neural information processing systems},
  volume={28},
  year={2015}
}

@book{hastie2009elements,
  title={The elements of statistical learning: data mining, inference, and prediction},
  author={Hastie, Trevor and Tibshirani, Robert and Friedman, Jerome H and Friedman, Jerome H},
  volume={2},
  year={2009},
  publisher={Springer}
}

@article{sebastiani2002machine,
  title={Machine learning in automated text categorization},
  author={Sebastiani, Fabrizio},
  journal={ACM computing surveys (CSUR)},
  volume={34},
  number={1},
  pages={1--47},
  year={2002},
  publisher={ACM New York, NY, USA}
}

@article{debnath1991structure,
  title={Structure-activity relationship of mutagenic aromatic and heteroaromatic nitro compounds. correlation with molecular orbital energies and hydrophobicity},
  author={Debnath, Asim Kumar and Lopez de Compadre, Rosa L and Debnath, Gargi and Shusterman, Alan J and Hansch, Corwin},
  journal={Journal of medicinal chemistry},
  volume={34},
  number={2},
  pages={786--797},
  year={1991},
  publisher={ACS Publications}
}

@article{morris2020tudataset,
  title={Tudataset: A collection of benchmark datasets for learning with graphs},
  author={Morris, Christopher and Kriege, Nils M and Bause, Franka and Kersting, Kristian and Mutzel, Petra and Neumann, Marion},
  journal={arXiv preprint arXiv:2007.08663},
  year={2020}
}

@article{toivonen2003statistical,
  title={Statistical evaluation of the predictive toxicology challenge 2000--2001},
  author={Toivonen, Hannu and Srinivasan, Ashwin and King, Ross D and Kramer, Stefan and Helma, Christoph},
  journal={Bioinformatics},
  volume={19},
  number={10},
  pages={1183--1193},
  year={2003},
  publisher={Oxford University Press}
}

@article{liu2015data,
  title={Data-driven identification of structural alerts for mitigating the risk of drug-induced human liver injuries},
  author={Liu, Ruifeng and Yu, Xueping and Wallqvist, Anders},
  journal={Journal of cheminformatics},
  volume={7},
  pages={1--8},
  year={2015},
  publisher={Springer}
}

@article{yang2019analyzing,
  title={Analyzing learned molecular representations for property prediction},
  author={Yang, Kevin and Swanson, Kyle and Jin, Wengong and Coley, Connor and Eiden, Philipp and Gao, Hua and Guzman-Perez, Angel and Hopper, Timothy and Kelley, Brian and Mathea, Miriam and others},
  journal={Journal of chemical information and modeling},
  volume={59},
  number={8},
  pages={3370--3388},
  year={2019},
  publisher={ACS Publications}
}

@article{xia2024understanding,
  title={Understanding the limitations of deep models for molecular property prediction: Insights and solutions},
  author={Xia, Jun and Zhang, Lecheng and Zhu, Xiao and Liu, Yue and Gao, Zhangyang and Hu, Bozhen and Tan, Cheng and Zheng, Jiangbin and Li, Siyuan and Li, Stan Z},
  journal={Advances in Neural Information Processing Systems},
  volume={36},
  year={2024}
}

@inproceedings{liu2022rethinking,
  title={Rethinking attention-model explainability through faithfulness violation test},
  author={Liu, Yibing and Li, Haoliang and Guo, Yangyang and Kong, Chenqi and Li, Jing and Wang, Shiqi},
  booktitle={International Conference on Machine Learning},
  pages={13807--13824},
  year={2022},
  organization={PMLR}
}

@article{achiam2023gpt,
  title={Gpt-4 technical report},
  author={Achiam, Josh and Adler, Steven and Agarwal, Sandhini and Ahmad, Lama and Akkaya, Ilge and Aleman, Florencia Leoni and Almeida, Diogo and Altenschmidt, Janko and Altman, Sam and Anadkat, Shyamal and others},
  journal={arXiv preprint arXiv:2303.08774},
  year={2023}
}

@article{dubey2024llama,
  title={The llama 3 herd of models},
  author={Dubey, Abhimanyu and Jauhri, Abhinav and Pandey, Abhinav and Kadian, Abhishek and Al-Dahle, Ahmad and Letman, Aiesha and Mathur, Akhil and Schelten, Alan and Yang, Amy and Fan, Angela and others},
  journal={arXiv preprint arXiv:2407.21783},
  year={2024}
}

@article{quinlan1986induction,
  title={Induction of decision trees},
  author={Quinlan, J. Ross},
  journal={Machine learning},
  volume={1},
  pages={81--106},
  year={1986},
  publisher={Springer}
}

@inproceedings{chen2016xgboost,
  title={Xgboost: A scalable tree boosting system},
  author={Chen, Tianqi and Guestrin, Carlos},
  booktitle={Proceedings of the 22nd acm sigkdd international conference on knowledge discovery and data mining},
  pages={785--794},
  year={2016}
}

@article{breiman2001random,
  title={Random forests},
  author={Breiman, Leo},
  journal={Machine learning},
  volume={45},
  pages={5--32},
  year={2001},
  publisher={Springer}
}

@article{irwin2012zinc,
  title={ZINC: a free tool to discover chemistry for biology},
  author={Irwin, John J and Sterling, Teague and Mysinger, Michael M and Bolstad, Erin S and Coleman, Ryan G},
  journal={Journal of chemical information and modeling},
  volume={52},
  number={7},
  pages={1757--1768},
  year={2012},
  publisher={ACS Publications}
}

@article{moiclr,
  title={Molecular contrastive learning of representations via graph neural networks},
  author={Wang, Yuyang and Wang, Jianren and Cao, Zhonglin and Barati Farimani, Amir},
  journal={Nature Machine Intelligence},
  volume={4},
  number={3},
  pages={279--287},
  year={2022},
  publisher={Nature Publishing Group UK London}
}

@article{cgib,
  title={Conditional Graph Information Bottleneck for Molecular Relational Learning. arXiv 2023},
  author={Lee, N and Hyun, D and Na, GS and Kim, S and Lee, J and Park, C},
  journal={arXiv preprint arXiv:2305.01520},
  year={2023}
}

@article{mose,
  title={Homomorphism Counts as Structural Encodings for Graph Learning},
  author={Bao, Linus and Jin, Emily and Bronstein, Michael and Ceylan, {\.I}smail {\.I}lkan and Lanzinger, Matthias},
  journal={International Conference on Learning Representations},
  year={2025}
}

@article{fragnet,
  title={FragNet: A Graph Neural Network for Molecular Property Prediction with Four Layers of Interpretability},
  author={Panapitiya, Gihan and Gao, Peiyuan and Maupin, C Mark and Saldanha, Emily G},
  journal={arXiv preprint arXiv:2410.12156},
  year={2024}
}

@inproceedings{momentgnn,
  title={Counting graph substructures with graph neural networks},
  author={Kanatsoulis, Charilaos and Ribeiro, Alejandro},
  booktitle={The twelfth international conference on learning representations},
  year={2024}
}

@article{s-cgib,
  title={Pre-training Graph Neural Networks on Molecules by Using Subgraph-Conditioned Graph Information Bottleneck},
  author={Hoang, Van Thuy and Lee, O and others},
  journal={arXiv preprint arXiv:2412.15589},
  year={2024}
}

@inproceedings{graphmvp,
  title={Pre-training Molecular Graph Representation with 3D Geometry},
  author={Liu, Shengchao and Wang, Hanchen and Liu, Weiyang and Lasenby, Joan and Guo, Hongyu and Tang, Jian},
  booktitle={ICLR 2022 Workshop on Geometrical and Topological Representation Learning},
  year={2022}
}

@article{wu2018moleculenet,
  title={MoleculeNet: a benchmark for molecular machine learning},
  author={Wu, Zhenqin and Ramsundar, Bharath and Feinberg, Evan N and Gomes, Joseph and Geniesse, Caleb and Pappu, Aneesh S and Leswing, Karl and Pande, Vijay},
  journal={Chemical science},
  volume={9},
  number={2},
  pages={513--530},
  year={2018},
  publisher={Royal Society of Chemistry}
}

@article{zheng2025large,
  title={Large language models for scientific discovery in molecular property prediction},
  author={Zheng, Yizhen and Koh, Huan Yee and Ju, Jiaxin and Nguyen, Anh TN and May, Lauren T and Webb, Geoffrey I and Pan, Shirui},
  journal={Nature Machine Intelligence},
  pages={1--11},
  year={2025},
  publisher={Nature Publishing Group UK London}
}

@article{fang2022geometry,
  title={Geometry-enhanced molecular representation learning for property prediction},
  author={Fang, Xiaomin and Liu, Lihang and Lei, Jieqiong and He, Donglong and Zhang, Shanzhuo and Zhou, Jingbo and Wang, Fan and Wu, Hua and Wang, Haifeng},
  journal={Nature Machine Intelligence},
  volume={4},
  number={2},
  pages={127--134},
  year={2022},
  publisher={Nature Publishing Group UK London}
}

@inproceedings{ren2024representation,
  title={Representation learning with large language models for recommendation},
  author={Ren, Xubin and Wei, Wei and Xia, Lianghao and Su, Lixin and Cheng, Suqi and Wang, Junfeng and Yin, Dawei and Huang, Chao},
  booktitle={Proceedings of the ACM Web Conference 2024},
  pages={3464--3475},
  year={2024}
}

@article{wellawatte2025human,
  title={Human interpretable structure-property relationships in chemistry using explainable machine learning and large language models},
  author={Wellawatte, Geemi P and Schwaller, Philippe},
  journal={Communications Chemistry},
  volume={8},
  number={1},
  pages={11},
  year={2025},
  publisher={Nature Publishing Group UK London}
}

@article{balaji2023gpt,
  title={Gpt-molberta: Gpt molecular features language model for molecular property prediction},
  author={Balaji, Suryanarayanan and Magar, Rishikesh and Jadhav, Yayati and Farimani, Amir Barati},
  journal={arXiv preprint arXiv:2310.03030},
  year={2023}
}

@book{bruice2017organic,
  title={Organic chemistry},
  author={Bruice, Paula Yurkanis},
  year={2017},
  publisher={Pearson}
}

@article{mcinnes2018umap,
  title={Umap: Uniform manifold approximation and projection for dimension reduction},
  author={McInnes, Leland and Healy, John and Melville, James},
  journal={arXiv preprint arXiv:1802.03426},
  year={2018}
}

@inproceedings{bai2021attentions,
  title={Why attentions may not be interpretable?},
  author={Bai, Bing and Liang, Jian and Zhang, Guanhua and Li, Hao and Bai, Kun and Wang, Fei},
  booktitle={Proceedings of the 27th ACM SIGKDD conference on knowledge discovery \& data mining},
  pages={25--34},
  year={2021}
}

@article{hase2020evaluating,
  title={Evaluating explainable AI: Which algorithmic explanations help users predict model behavior?},
  author={Hase, Peter and Bansal, Mohit},
  journal={arXiv preprint arXiv:2005.01831},
  year={2020}
}

@article{kunz2024properties,
  title={Properties and challenges of llm-generated explanations},
  author={Kunz, Jenny and Kuhlmann, Marco},
  journal={arXiv preprint arXiv:2402.10532},
  year={2024}
}

@article{zheng2024large,
  title={Large language models in drug discovery and development: From disease mechanisms to clinical trials},
  author={Zheng, Yizhen and Koh, Huan Yee and Yang, Maddie and Li, Li and May, Lauren T and Webb, Geoffrey I and Pan, Shirui and Church, George},
  journal={arXiv preprint arXiv:2409.04481},
  year={2024}
}

@article{petroni2019language,
  title={Language models as knowledge bases?},
  author={Petroni, Fabio and Rockt{\"a}schel, Tim and Lewis, Patrick and Bakhtin, Anton and Wu, Yuxiang and Miller, Alexander H and Riedel, Sebastian},
  journal={arXiv preprint arXiv:1909.01066},
  year={2019}
}

@inproceedings{
yang2025curriculumaware,
title={Curriculum-aware Training for Discriminating Molecular Property Prediction Models},
author={Hansi Yang and Quanming Yao and James Kwok},
booktitle={The Thirteenth International Conference on Learning Representations},
year={2025}
}

@inproceedings{
zhou2025clmfap,
title={{CL}-{MFAP}: A Contrastive Learning-Based Multimodal Foundation Model for Molecular Property Prediction and Antibiotic Screening},
author={Gen Zhou and Sugitha Janarthanan and Yutong Lu and Pingzhao Hu},
booktitle={The Thirteenth International Conference on Learning Representations},
year={2025}
}

@inproceedings{
zhong2025automatic,
title={Automatic Auxiliary Task Selection and Adaptive Weighting Boost Molecular Property Prediction},
author={Zhiqiang Zhong and Davide Mottin},
booktitle={The Thirty-ninth Annual Conference on Neural Information Processing Systems},
year={2025}
}

@inproceedings{
yeyunchen2025mipt,
title={{MIPT}: Multilevel Informed Prompt Tuning for Robust Molecular Property Prediction},
author={yeyunchen and Jiangming Shi},
booktitle={Forty-second International Conference on Machine Learning},
year={2025}
}

@inproceedings{
rao2025quadruple,
title={Quadruple Attention in Many-body Systems for Accurate Molecular Property Predictions},
author={Jiahua Rao and Dahao Xu and Wentao Wei and Yicong Chen and Mingjun Yang and Yuedong Yang},
booktitle={Forty-second International Conference on Machine Learning},
year={2025}
}

@inproceedings{
li2024towards,
title={Towards 3D Molecule-Text Interpretation in Language Models},
author={Sihang Li and Zhiyuan Liu and Yanchen Luo and Xiang Wang and Xiangnan He and Kenji Kawaguchi and Tat-Seng Chua and Qi Tian},
booktitle={The Twelfth International Conference on Learning Representations},
year={2024}
}

@article{feng2024generation,
  title={Generation of 3D molecules in pockets via a language model},
  author={Feng, Wei and Wang, Lvwei and Lin, Zaiyun and Zhu, Yanhao and Wang, Han and Dong, Jianqiang and Bai, Rong and Wang, Huting and Zhou, Jielong and Peng, Wei and others},
  journal={Nature Machine Intelligence},
  volume={6},
  number={1},
  pages={62--73},
  year={2024}
}

@inproceedings{
li2025geometry,
title={Geometry Informed Tokenization of Molecules for Language Model Generation},
author={Xiner Li and Limei Wang and Youzhi Luo and Carl Edwards and Shurui Gui and Yuchao Lin and Heng Ji and Shuiwang Ji},
booktitle={Forty-second International Conference on Machine Learning},
year={2025}
}

@inproceedings{
liu2025multimodal,
title={Multimodal Large Language Models for Inverse Molecular Design with Retrosynthetic Planning},
author={Gang Liu and Michael Sun and Wojciech Matusik and Meng Jiang and Jie Chen},
booktitle={The Thirteenth International Conference on Learning Representations},
year={2025}
}

@article{kang2024quantitative,
  title={A quantitative and qualitative evaluation of LLM-based explainable fault localization},
  author={Kang, Sungmin and An, Gabin and Yoo, Shin},
  journal={Proceedings of the ACM on Software Engineering},
  volume={1},
  number={FSE},
  pages={1424--1446},
  year={2024}
}

@article{sharkey2025open,
  title={Open problems in mechanistic interpretability},
  author={Sharkey, Lee and Chughtai, Bilal and Batson, Joshua and Lindsey, Jack and Wu, Jeff and Bushnaq, Lucius and Goldowsky-Dill, Nicholas and Heimersheim, Stefan and Ortega, Alejandro and Bloom, Joseph and others},
  journal={arXiv preprint arXiv:2501.16496},
  year={2025}
}

@article{modell2025origins,
  title={The Origins of Representation Manifolds in Large Language Models},
  author={Modell, Alexander and Rubin-Delanchy, Patrick and Whiteley, Nick},
  journal={arXiv preprint arXiv:2505.18235},
  year={2025}
}

@inproceedings{
park2024the,
title={The Linear Representation Hypothesis and the Geometry of Large Language Models},
author={Kiho Park and Yo Joong Choe and Victor Veitch},
booktitle={Forty-first International Conference on Machine Learning},
year={2024}
}

@inproceedings{
he2024harnessing,
title={Harnessing Explanations: {LLM}-to-{LM} Interpreter for Enhanced Text-Attributed Graph Representation Learning},
author={Xiaoxin He and Xavier Bresson and Thomas Laurent and Adam Perold and Yann LeCun and Bryan Hooi},
booktitle={The Twelfth International Conference on Learning Representations},
year={2024}
}

@article{bader1994theoretical,
  title={Theoretical definition of a functional group and the molecular orbital paradigm},
  author={Bader, Richard Frederick William and Popelier, Paul Lode Albert and Keith, Todd Alan},
  journal={Angewandte Chemie International Edition in English},
  volume={33},
  number={6},
  pages={620--631},
  year={1994}
}

@article{abdi2010principal,
  title={Principal component analysis},
  author={Abdi, Herv{\'e} and Williams, Lynne J},
  journal={Wiley interdisciplinary reviews: computational statistics},
  volume={2},
  number={4},
  pages={433--459},
  year={2010},
  publisher={Wiley Online Library}
}

@incollection{rao1995linear,
  title={Linear models},
  author={Rao, Calyampudi Radhakrishna and Toutenburg, Helge},
  booktitle={Linear Models: Least Squares and Alternatives},
  pages={3--18},
  year={1995},
  publisher={Springer}
}

@article{siddique2024exploring,
  title={Exploring functional groups and molecular structures: a comprehensive analysis using FTIR spectroscopy},
  author={Siddique, Iqtiar},
  journal={Chemistry Research Journal},
  volume={9},
  number={2},
  pages={70--76},
  year={2024}
}

@article{gani2019group,
  title={Group contribution-based property estimation methods: advances and perspectives},
  author={Gani, Rafiqul},
  journal={Current Opinion in Chemical Engineering},
  volume={23},
  pages={184--196},
  year={2019},
  publisher={Elsevier}
}

@article{yan2010oxidation,
  title={Oxidation functional groups on graphene: Structural and electronic properties},
  author={Yan, Jia-An and Chou, MY},
  journal={Physical Review B—Condensed Matter and Materials Physics},
  volume={82},
  number={12},
  pages={125403},
  year={2010},
  publisher={APS}
}

@book{murphy2012machine,
  title={Machine learning: a probabilistic perspective},
  author={Murphy, Kevin P},
  year={2012},
  publisher={MIT press}
}

@article{poole1971assumptions,
  title={The assumptions of the linear regression model},
  author={Poole, Michael A and O'Farrell, Patrick N},
  journal={Transactions of the Institute of British Geographers},
  pages={145--158},
  year={1971},
  publisher={JSTOR}
}

@article{schmidt2018linear,
  title={Linear regression and the normality assumption},
  author={Schmidt, Amand F and Finan, Chris},
  journal={Journal of clinical epidemiology},
  volume={98},
  pages={146--151},
  year={2018},
  publisher={Elsevier}
}

@article{hastie2015statistical,
  title={Statistical learning with sparsity},
  author={Hastie, Trevor and Tibshirani, Robert and Wainwright, Martin},
  journal={Monographs on statistics and applied probability},
  volume={143},
  number={143},
  pages={8},
  year={2015}
}

@article{kaplan2020scaling,
  title={Scaling laws for neural language models},
  author={Kaplan, Jared and McCandlish, Sam and Henighan, Tom and Brown, Tom B and Chess, Benjamin and Child, Rewon and Gray, Scott and Radford, Alec and Wu, Jeffrey and Amodei, Dario},
  journal={arXiv preprint arXiv:2001.08361},
  year={2020}
}
\bibliographystyle{IEEEtran}


\end{document}